\documentclass[runningheads]{llncs}
\usepackage{graphicx}

\usepackage{import}
\usepackage{amsmath,amssymb} 
\usepackage{tikz}
\usepackage{comment}
\usepackage{color}
\usepackage{orcidlink} 
\usepackage[accsupp]{axessibility}  

\usepackage{biblatex} 
\usepackage{multirow}
\usepackage[export]{adjustbox}
\usepackage{wrapfig}
\usepackage[export]{adjustbox}

\usepackage{cleveref}
\usepackage{hyperref} 
\Crefformat{figure}{#2Fig.~#1#3}
\Crefmultiformat{figure}{Figs.~#2#1#3}{ and~#2#1#3}{, #2#1#3}{ and~#2#1#3}
\usepackage{booktabs}

\usepackage{hyperref}
\hypersetup{
  colorlinks=false,
  linkbordercolor=white,
 urlbordercolor=white,
pdfborder={0 0 0}
} 

\begin{document}

\pagestyle{headings}
\mainmatter
\def\ECCVSubNumber{7476}  

\title{DeepShadow: Neural Shape from Shadow}

\titlerunning{DeepShadow: Neural Shape from Shadow}

\author{Asaf Karnieli$^{\orcidlink{0000-0003-3562-7463}}$ \and
Ohad Fried$^{\orcidlink{0000-0001-7109-4006}}$ \and
Yacov Hel-Or$^{\orcidlink{0000-0002-6880-3374}}$}
\authorrunning{A. Karnieli et al.}
%
\institute{School of Computer Science, Reichman University \\
\email{asafkarnieli@gmail.com}\\
\email{\{ofried,toky\}@runi.ac.il}}


\maketitle

\def\ShowNotes{}
\newcommand{\betweencellpdf}{\ensuremath{h^c}}
\newcommand{\betweencellpdffine}{\ensuremath{h^f}}
\newcommand{\approxbetweencellpdffine}{\ensuremath{\hat{h}^f}}
\newcommand{\incellpdf}{\ensuremath{f}}
\newcommand{\incellpdfnormalized}{\ensuremath{f'}}

\newcommand{\incellcdf}{\ensuremath{F}}

\newcommand{\totalpdf}{\ensuremath{f_{dd}}}
\newcommand{\totalcdf}{\ensuremath{F_{dd}}}

\newcommand{\ignorethis}[1]{}
\newcommand{\redund}[1]{#1}

\newcommand{\apriori    }     {\textit{a~priori}}
\newcommand{\aposteriori}     {\textit{a~posteriori}}
\newcommand{\perse      }     {\textit{per~se}}
\newcommand{\naive      }     {{na\"{\i}ve}}
\newcommand{\Naive      }     {{Na\"{\i}ve}}
\newcommand{\Identity   }     {\mat{I}}
\newcommand{\Zero       }     {\mathbf{0}}
\newcommand{\Reals      }     {{\textrm{I\kern-0.18em R}}}
\newcommand{\isdefined  }     {\mbox{\hspace{0.5ex}:=\hspace{0.5ex}}}
\newcommand{\texthalf   }     {\ensuremath{\textstyle\frac{1}{2}}}
\newcommand{\half       }     {\ensuremath{\frac{1}{2}}}
\newcommand{\third      }     {\ensuremath{\frac{1}{3}}}
\newcommand{\fourth     }     {\ensuremath{\frac{1}{4}}}

\newcommand{\Lone} {\ensuremath{L_1}}
\newcommand{\Ltwo} {\ensuremath{L_2}}

\newcommand{\degree} {\ensuremath{^{\circ}}}

\newcommand{\mat        } [1] {{\text{\boldmath $\mathbit{#1}$}}}
\newcommand{\Approx     } [1] {\widetilde{#1}}
\newcommand{\change     } [1] {\mbox{{\footnotesize $\Delta$} \kern-3pt}#1}

\newcommand{\Order      } [1] {O(#1)}
\newcommand{\set        } [1] {{\lbrace #1 \rbrace}}
\newcommand{\floor      } [1] {{\lfloor #1 \rfloor}}
\newcommand{\ceil       } [1] {{\lceil  #1 \rceil }}
\newcommand{\inverse    } [1] {{#1}^{-1}}
\newcommand{\transpose  } [1] {{#1}^\mathrm{T}}
\newcommand{\invtransp  } [1] {{#1}^{-\mathrm{T}}}
\newcommand{\relu       } [1] {{\lbrack #1 \rbrack_+}}

\newcommand{\abs        } [1] {{| #1 |}}
\newcommand{\Abs        } [1] {{\left| #1 \right|}}
\newcommand{\norm       } [1] {{\| #1 \|}}
\newcommand{\Norm       } [1] {{\left\| #1 \right\|}}
\newcommand{\pnorm      } [2] {\norm{#1}_{#2}}
\newcommand{\Pnorm      } [2] {\Norm{#1}_{#2}}
\newcommand{\inner      } [2] {{\langle {#1} \, | \, {#2} \rangle}}
\newcommand{\Inner      } [2] {{\left\langle \begin{array}{@{}c|c@{}}
                               \displaystyle {#1} & \displaystyle {#2}
                               \end{array} \right\rangle}}

\newcommand{\twopartdef}[4]
{
  \left\{
  \begin{array}{ll}
    #1 & \mbox{if } #2 \\
    #3 & \mbox{if } #4
  \end{array}
  \right.
}

\newcommand{\fourpartdef}[8]
{
  \left\{
  \begin{array}{ll}
    #1 & \mbox{if } #2 \\
    #3 & \mbox{if } #4 \\
    #5 & \mbox{if } #6 \\
    #7 & \mbox{if } #8
  \end{array}
  \right.
}

\newcommand{\len}[1]{\text{len}(#1)}

\newlength{\w}
\newlength{\h}
\newlength{\x}

\definecolor{darkred}{rgb}{0.7,0.1,0.1}
\definecolor{darkgreen}{rgb}{0.1,0.6,0.1}
\definecolor{cyan}{rgb}{0.7,0.0,0.7}
\definecolor{otherblue}{rgb}{0.1,0.4,0.8}
\definecolor{maroon}{rgb}{0.76,.13,.28}
\definecolor{burntorange}{rgb}{0.81,.33,0}

\ifdefined\ShowNotes
  \newcommand{\colornote}[3]{{\color{#1}\textbf{#2} #3\normalfont}}
\else
  \newcommand{\colornote}[3]{}
\fi

\newcommand {\todo}[1]{\colornote{cyan}{TODO}{#1}}
\newcommand {\ohad}[1]{\colornote{burntorange}{OF:}{#1}}
\newcommand {\toky}[1]{\colornote{darkgreen}{YH:}{#1}}
\newcommand {\asaf}[1]{\colornote{otherblue}{AK:}{#1}}

\newcommand {\reqs}[1]{\colornote{red}{\tiny #1}}

\newcommand {\new}[1]{\colornote{red}{#1}}

\newcommand*\rot[1]{\rotatebox{90}{#1}}

\newcommand {\newstuff}[1]{#1}

\newcommand\todosilent[1]{}

\newcommand{\woBGmask}{{w/o~bg~\&~mask}}
\newcommand{\woMask}{{w/o~mask}}

\providecommand{\keywords}[1]
{
  \textbf{\textit{Keywords---}} #1
}

\newcommand {\shortcite}[1]{\cite{#1}}

\begin{figure}[h!]
\centering
\includegraphics[width=0.9\textwidth]{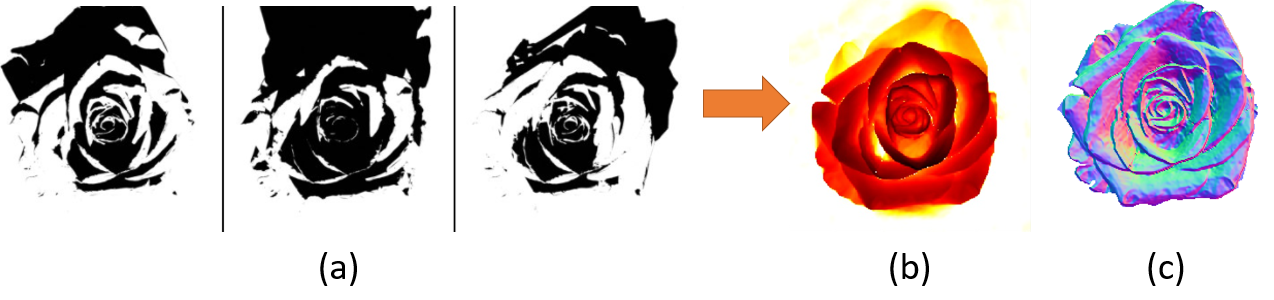}
\caption[Results on Rose Dataset] 
{Results on the rose object. (a) Input shadow maps, which were used for supervision 
(b) Depth produced by the algorithm. (c) Surface normals extracted from the depth map.}
\label{fig:banner}
\end{figure}

\begin{abstract}

This paper presents `DeepShadow', a one-shot method for recovering the depth map 
and surface normals from photometric stereo shadow maps. 
Previous works that try to recover the surface normals from 
photometric stereo images treat cast shadows as a disturbance. 
We show that the self and cast shadows not only do not disturb 3D reconstruction, 
but can be used alone, as a strong learning signal, 
to recover the depth map and surface normals.
We demonstrate that 3D reconstruction from shadows
can even outperform shape-from-shading in certain cases.
To the best of our knowledge, our method is the first to reconstruct 3D shape-from-shadows
using neural networks. The method does not require any
pre-training or expensive labeled data, and is optimized during inference time.

\end{abstract}

\section{Introduction}

The photometric stereo setting was first defined in \cite{10.1117/12.7972479}. 
The setting includes a single camera viewpoint, and multiple varying illumination sources.
Most works try to extract the underlying per-pixel normal map from the input images.
The original problem assumed Lambertian objects, 
which only have a diffuse reflectance component.
More recent works solve a more general setting, 
with various light and material properties, including specular components.

To date, most works extract the required information from the local illumination effect, 
defined by the bidirectional reflectance distribution function (BRDF),
and either ignored the global cast shadow effect, or model it in a statistical way.
Almost all recent works use Conv Nets to recover the 3D structure 
from photometric stereo. Since Conv Nets have limited receptive 
fields, global information can only be aggregated in the deeper layers where accurate 
spatial resolution is limited.

In our work, we extract the depth information directly from the cast shadows.
As far as we know, this is the first attempt to do so using neural networks.
In the photometric stereo setting, cast and attached shadow detection is a relatively 
easy learning task. The shadow maps are binary inputs, i.e., have a 0 or 1 value.
To extract depth, we initially predict per-pixel depth information in 2D 
and thereafter produce a point-cloud in 3D from this prediction.
The 3D points are then used to calculate global cast shadows, by tracing the light
source to all destination pixels. The input images are used as supervision for
the produced shadow maps.
The entire process is differentiable, and the predicted depth can be learned using optimization.
Our process is physics based and can aggregate global information while keeping 
the full resolution of the image.

Most previous works assume directional lights due to its simplicity, 
although this is usually not the case.
Many scenes consist of point light sources, thus depth information should be extracted
using a point light shading model.
In this work, we assume images generated by point light sources, 
but we can easily extend our method to handle directional lights.
Although the inputs in this work are only shadow maps, 
our method can be integrated with existing shape-from-shading models, 
in order to improve their results in complex shadowed scenes.
Our code and data are available at
\href{https://asafkar.github.io/deepshadow/}{https://asafkar.github.io/deepshadow/}.





\section{Related Work}
\label{sec:related}

\textbf{Shape Reconstruction from Shadows}
Extracting shape information from shadows has been attempted in various works.
This was mostly done before the deep learning era.
An early attempt was performed in \cite{698646}, which initializes a shadow map and
sets lower and upper bounds on expected illuminated or shadowed pixels, 
which are then optimized until reaching a predicted height map.
In \cite{shadow_graphs}, a graph-based representation is used to incorporate constraints
induced by the shadow maps. This work also optimizes the height map using iteration
based on low and high bounds.
The ShadowCuts \cite{shadow_cuts} work performs Lambertian photometric stereo with 
shadows present. It initially uses a graph-cut method for estimating light source visibility
for every pixel. Then it uses the produced shadow maps along with shading information to
perform surface normal integration, to produce the final predicted surface.
In \cite{recovering3d_shape_and_light}, the authors use 1D shadow graphs  to speed
up the 3D reconstruction process. 
1D slices (i.e., rows or columns of the shadow images) are 
used, assuming the light source moves in two independent circular orbits. 
Each such slice can be solved independently, 
to retrieve the corresponding 1D slice of the underlying height map.
This method only handles light sources that are on a unit sphere, and the trajectories
of the light sources must be perpendicular to each other.
All the above works extract the shadow maps from the images using a hand-picked threshold.
A recent method \cite{Lyu_2021_ICCV} uses an initial approximation of the 3D geometry as input to optimize
object deformations by using soft shadow information. In contrary, our method does not require an initial
approximation of the geometry.
The method uses spherical harmonics to approximate the surface occlusion whereas we use implicit representations 
along with a linear-time tracing method.

\textbf{Implicit Representations} 
Implicit representations have recently been used to replace voxel grids or 
meshes with continues parameterization using neural networks. 
These are used in \cite{mildenhall2020nerf}
to query the color and opacity of a certain point in space, to produce the pixel
color of an image acquired from a specific viewing position.
Lately, implicit representations have also been used in works such as
\cite{boss2021nerd,srinivasan2020nerv,2021} to recover various underlying geometric 
and photometric properties such as albedo, normals, etc.
These have all been done in multi-view settings and take many hours and days
to optimize.

\textbf{Viewshed Analysis} Viewshed analysis solves the problem of which areas are
visible from a specific viewpoint, and solved in
\cite{cole1989visibility,goodchild1989coverage} using a line-of-sight (LOS) algorithm.
This algorithm tracks along a ray from the viewpoint to the target point, while 
verifying that the target point is not occluded by the height points along the ray.
Shapira \cite{shapira1990visibility} proposed the R3 method for viewshed computation, 
by generating LOS rays from the viewpoint to all other cells (pixels) in a height map.
Franklin et al. \cite{franklin1994higher} introduced the R2 algorithm, which optimizes
the R3 method by launching LOS rays only to the outer cells (boundary pixels),
while computing and storing
the intermediate results of all cells on the way. This method is considered an approximation
since some rays will encounter the same cells during the process.

\textbf{Photometric Stereo} In the past few years, various learning-based methods have
been proposed to solve calibrated and uncalibrated photometric stereo problems
\cite{kaya2021uncalibrated,chen2019selfcalibrating,chen2020deep, yao2020gps,logothetis2021pxnet}.
Most existing methods solve the problem in a supervised manner, 
i.e., given inputs of N images and their associated light directions and intensities, 
the methods predict the surface normals as 
an output. The ground-truth normals are needed as supervision in order to train the network.
In the uncalibrated scenarios, the light intensities and directions are not required. 
Works such as \cite{chen2019selfcalibrating,10.1007/978-3-030-58568-6_44} first regress the 
light direction and light intensities, 
and then solve the calibrated photometric stereo problem
using the predicted lights.
Methods such as \cite{kaya2021uncalibrated,taniai2018neural} solve
the problem in an unsupervised manner, using a reconstruction loss. 
These methods only require the input images, which are used as self-supervision in the
reconstruction loss.  

Most of the above-mentioned methods use the pixelwise information to solve the problem.
No inter-pixel relations are taken into account, besides the obvious local dependencies
resulting from the local filtering.
There are also methods such as \cite{yao2020gps,logothetis2021pxnet} that use
local neighborhood pixels as well by feature aggregation. 
In \cite{yao2020gps}, the authors use a graph-based neural network to unify 
all-pixel and per-pixel information, while in \cite{logothetis2021pxnet} the authors
use each pixel as a data-point to learn interactions between pixels. 
Although the latter method does handle cast
shadows, it does so within a bound area of the observation map and in a statistical manner.
Apart from this method, the previous methods ignore the cast shadows as a source of 
information, and treat them as a disturbance.

Some work solve the near-field photometric stereo problem,
which assumes that the lights are close to the object, 
and thus cannot use directional lights.
In \cite{DeepNearFieldPS}, the authors use a hybrid approach that utilizes
both distant and near-field
models. The surface normals of a local area are computed by assuming they are illuminated
by a distant light source, and the reconstruction error is based on the near field model.
This method yields good results for close range lights, 
although it requires knowing the mean depth for
initialization.
It also requires the location of the point lights as well as the ambient lighting.
The work in \cite{NearFieldCNN} also uses a far and near hybrid method, 
by first predicting the surface normals assuming far light sources, 
and then computing the depth
by integrating the normal fields.
The normal fields are then used to estimate light directions and 
attenuation that are used to compute the reflectance. 

We also mention \cite{peng2017iccvw, haefner2018pdsr}, which produce high resolution depth maps 
by using low resolution depth maps and photometric stereo images.
Since few photometric stereo-based methods target depth outputs, 
we will compare our work with \cite{peng2017iccvw}.

In summary, most previous shape-from-shading methods need a large amount of data for
training. Also, they ignore the cast shadow instead of using it as a source of information.
The main contributions of our method are: 
\begin{itemize}
  \item We propose the first deep-learning based method for recovering depth and
  surface  normals from shadow maps.
  \item Our method includes a shadow calculation component, 
  which globally aggregates information in the spatial space.
  \item Our method uses linear-time calculation rather than quadratic complexity in 
  \cite{mildenhall2020nerf} and its follow-up works. We also use 2D parameterization for 
  depth maps, avoiding the expensive 3D parameterization in NeRF like works.
  \item In contrast to most shape-from-shading methods, which use photometric stereo data, 
  our method is  insensitive to non-diffuse reflectance (specular highlights) or varying 
  intensity lights, which also enables it to generate good results in 
  both from the near-field and far-light photometric stereo settings.
  \item Lastly, our method estimates the depth-map in a one-shot manner, avoiding 
  costly data collection and supervised training.
\end{itemize}

\section{Shape-from-Shadow Method}

DeepShadow is a technique that estimates the shape of an object
or a scene from cast and attached shadows. 
The input data are multiple shadow maps and the location of their
associated point light sources. 
In our work, these can later be relaxed to input images alone.
Similarly to photometric stereo, the data are generated or captured with a single
viewpoint and multiple illuminations. 
We assume each input shadow map is a byproduct of a single image taken under a single light source.
In contrast to other methods that use directional lights, 
we do not enforce the location of the point light to be on a unit sphere.
Our algorithm can also work with directional light,
which models a light source at infinity.

\begin{figure}[t]
\includegraphics[width=\textwidth]{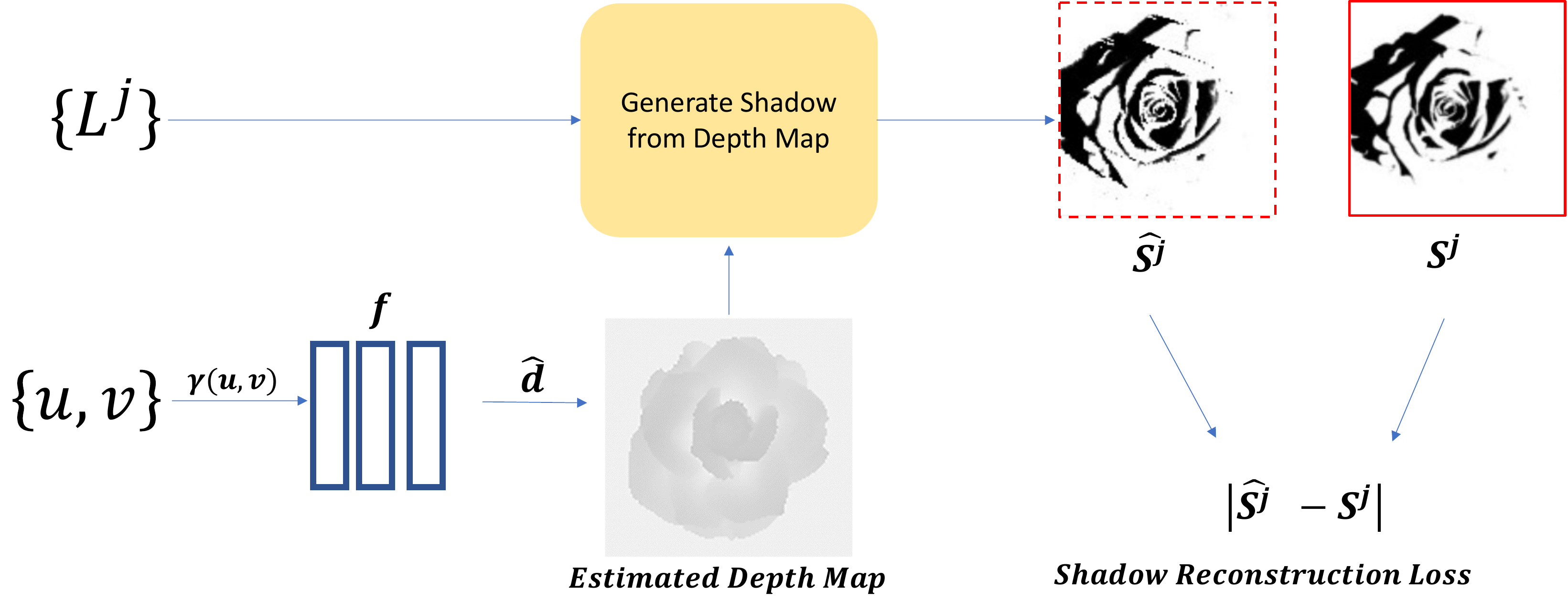}
\caption[Algorithm Overview]
{Algorithm Overview. DeepShadow takes the light source location $L^j$ and pixel 
coordinates $({u, v)}$ as inputs, along with the estimated depth $\hat{d}$ from the MLP, 
and outputs an estimate of the shadow map $\hat{S^j}$ at each pixel location.
The ground-truth shadow map $S^j$ is then used as a supervision to optimize the learned
depth map.}
\label{fig:algorithm_overview}
\end{figure}

\Cref{fig:algorithm_overview} shows our framework.
First, a multi-layer perceptron (MLP) predicts the depth at each given location $\mathbf{u}=(u,v)$
in the image. Positional encoding $\gamma(\mathbf{u})$ is used at the input of the MLP.  
Then, the predicted depth $\hat{d}$ along with an input light location $L^j$
are used to estimate a shadow map $\hat{S^j}$. 
This is a physics based component which is differentiable but not trainable.
The associated ground-truth shadow map $S^j$ is used for supervision to optimize the MLP.

\subsection{Shadow Map Estimation} 
To calculate the shadow of a pixel with respect to
a specific light source $L^j$, the `shadow line scan' algorithm is used.
This algorithm traces a light ray from the light source to the destination pixel, 
and determines whether the pixel is illuminated or shadowed. 
More details are provided in Section 3.2.

To produce a shadow map from an estimated depth map, we use both world
coordinates\footnotemark{}
$\mathbf{X}=\{x,y,z\}$ and image coordinates\footnotemark[\value{footnote}]
$\mathbf{u}=\{u,v\}$ on the image plane.
We assume the calibrated camera model is known. 
\footnotetext{We omit the homogeneous coordinates for the sake of clarity.}
During the process, we use perspective 
projection and unprojection to go from one coordinate system to the other.
Thus for a certain pixel location $\mathbf{u}_i$, $\mathbf{u}_i=P\mathbf{X}_i$, where
$P$ is the projection matrix $P=K\cdot[R|\mathbf{t}]$, $K$ is the intrinsic matrix and
$[R|\mathbf{t}]$ is the extrinsic matrix. 

For a given light source $L^j=(L_x^j,L_y^j,L_z^j)$ in world coordinates,
we initially map the light source to a point $\boldsymbol\ell^j=P \cdot L^j$ 
on the image plane.
To estimate the shadow for a chosen pixel $\mathbf{u}_i=(u_i, v_i)$ in the image,
a line $\mathbf{r}_i^j$ of points in the image plane is
generated between $ \boldsymbol \ell^j$
and $\mathbf{u}_i$:

\begin{equation}\label{eq:line_from_light_to_pixel_in_uv}
\mathbf{r}_i^j(\alpha)=(1-\alpha) \boldsymbol\ell^j + \alpha \mathbf{u}_i 
,~~ \alpha \in[0,1]
\end{equation}

\begin{figure}[t]
\includegraphics[width=\textwidth]{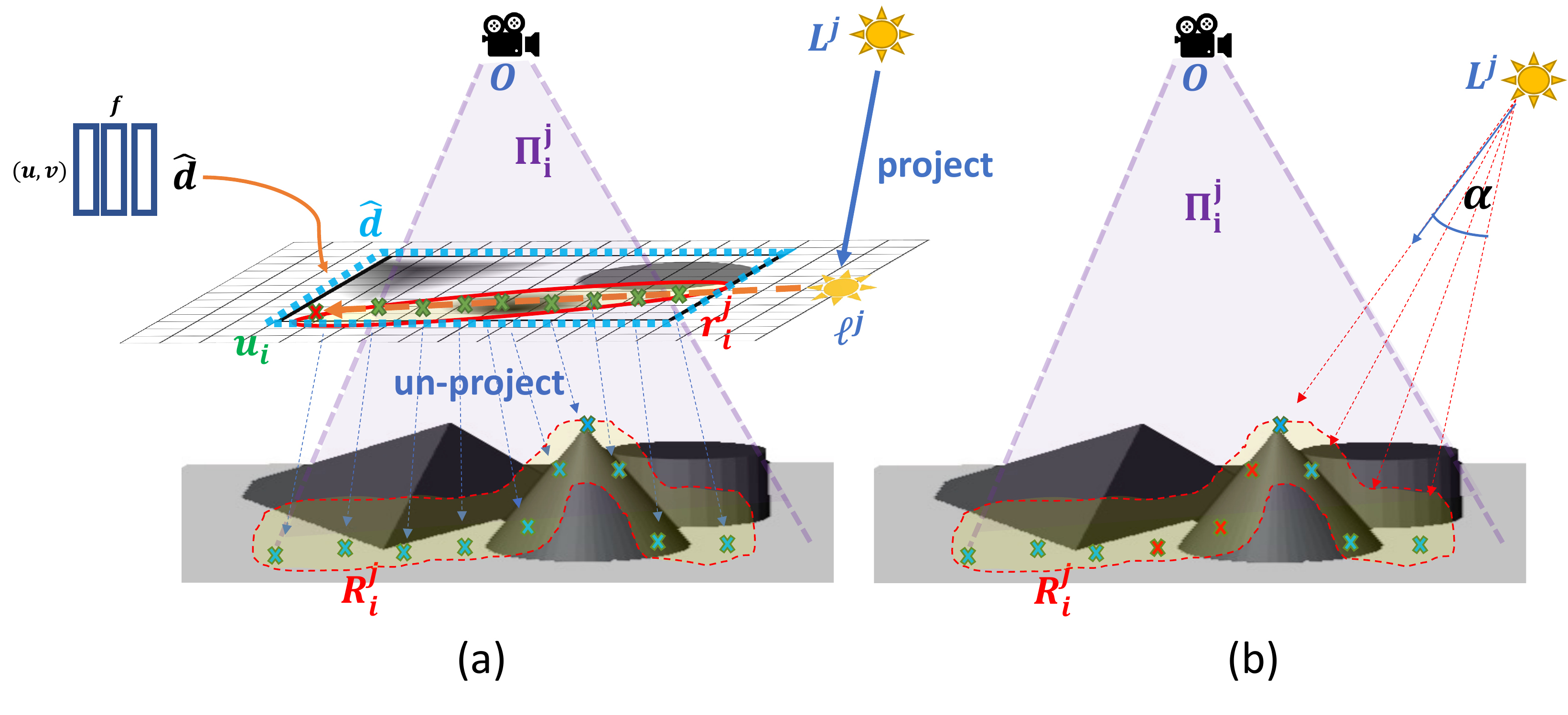}
\caption[Flow of Shape from Shadow]
{Flow of shape-from-shadow. 
(a) The light source $L^j$ is projected onto the image plane to receive $\boldsymbol\ell^j$.
A ray $\mathbf{r}_i^j$ of $(u,v)$ points is created between $\boldsymbol\ell^j$ and $\mathbf{u}_i$.
Then, each point with its estimated depth $\hat{d}$ is unprojected to world coordinates.
(b) The shadow line scan algorithm is used on points in 3D space to calculate
shadowed pixels. Red points are shadowed, since their angle to the light source
is smaller than $\alpha$.}
\label{fig:algo_flow}
\end{figure}

Points on $\mathbf{r}^j_i$ that are outside the image frame are excluded.
For each pixel in $\mathbf{r}_i^j(\alpha)$,
we estimate the depth $\hat{d}_i^j(\alpha)$ 
by querying the MLP model at the pixel's coordinates.
Each such triplet $(\mathbf{r}_i^j(\alpha)[x],\mathbf{r}_i^j(\alpha)[y],
\hat{d}_i^j(\alpha))$ is then unprojected to world coordinates 
to receive its 3D location $R_i^j(\alpha)$ 
as described in \Cref{eq:line_in_world_coords} and illustrated in \Cref{fig:algo_flow}a.

\begin{equation}\label{eq:line_in_world_coords}
R_i^j(\alpha)= P^{-1} \mathbf{r}_i^j(\alpha) \cdot \hat{d}_i^j(\alpha) 
~,~ \alpha \in[0,1]
\end{equation}

Once the 3D coordinates for each pixel in $R^j_i$ are obtained, 
we can solve this as a 1D line-of-sight problem, as illustrated in \Cref{fig:algo_flow}b.
Note, that this process will determine, \textit{for each} point in $R^j_i$,
whether it is shadowed or illuminated (and not only for the point $\mathbf{u}_i$).
We use the image plane coordinates for parameterization of the depth, 
thus avoiding a costly full 3D 
parameterization (as in \cite{mildenhall2020nerf} and follow up works).
This process assumes the object's depth is a function of $(u,v)$, 
which is a justifiable assumption since we are solving for the case of a single viewpoint.

Note that the light source $L^j$, viewer location $O$ and a given point $\mathbf{u}_i$
create a plane $\Pi_i^j$ in 3D, as can be seen in \Cref{fig:algo_flow}b. 
$\boldsymbol{\ell} ^j$ is also located
on $\Pi_i^j$, as are all the points in $\mathbf{r}_i^j$ and all the points in $R_j^j$.
This enables us to use a 1D line-of-sight algorithm between $L^j$ and all
points in $R_i^j$ (elaborated in Section 3.2).

We emphasize that the shadow calculation depends only on the
locations of the light source and the depth map. 
The camera's center of projection (COP) $O$ is used to generate the scan order 
for points along a 1D ray, and to expedite calculations as will be described 
in the next section.

\subsection{Shadow Line Scan Algorithm}

Given a height map and a viewer location, one can analyze all points visible to the viewer 
using a `line-of-sight' calculation. 
This process identifies which areas are visible from a given point. 
It can be naively achieved by sending 1D rays to every direction from the viewer, and 
calculating the line-of-sight visibility for each ray. 
This is analogous to producing shade maps -- the viewer
is replaced with a light source $L^j$,
and every pixel is analyzed to determine whether it is shadowed (visible from $L^j$) or not,
in the same manner.
We refer to this process as `shadow line scan'.


Calculating the visibility of each pixel point is time consuming and may take
many hours to train. Instead, we propose to calculate the visibility of entire lines in the
image. Each line $R^j_i$ is calculated using a single scan.
Each line includes the projections of all the points in 3D that are in the 
plane $\Pi^j_i$ that is formed by the COP $O$, $L^j$ and $\mathbf{u}_i$.

\begin{figure}[t]
\centering
\includegraphics[width=0.8\textwidth]{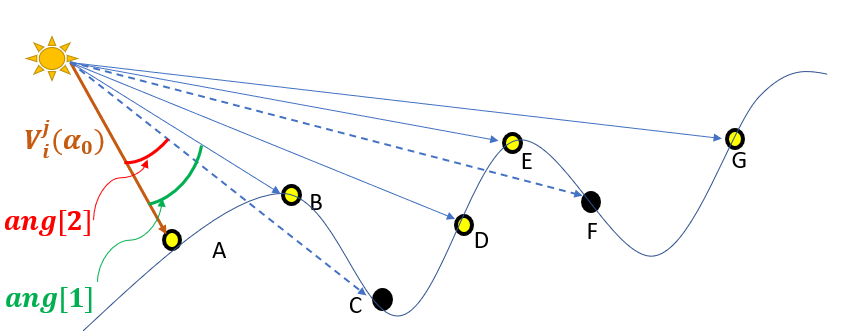}
\caption[.]
{1D Shadow Line Scan Algorithm for Shadow Calculation. 
Points A--B are visible to the viewer, thus have no shadows.
The angle $ang[2]$ between A and C (red) is smaller than the angle $ang[1]$ 
between A and B (green), thus C is not visible. 
Similarly, the angle between A and F is smaller than the last encountered largest angle, 
A--E, making F shadowed as well.}
\label{fig:viewshed}
\end{figure}

As is illustrated in \Cref{fig:viewshed}, we define a vector from the light source $L^j$ 
to each point in $R^j_i$, as 
\begin{equation}\label{eq:vector_light_to_points}
V_i^j(\alpha) = L^j-R_i^j(a).
\end{equation}

Since we use the shadow maps as supervision and the shadow maps are composed of discrete pixels,
we use discrete alpha values $\alpha \rightarrow{} \{\alpha_i \}^T_{i=0}$.
We calculate all the angles between $V_i^j[a_0]$ and each of the points.
The angle for the $i^{th}$ point is defined as

\begin{equation}
ang[i] = \arccos{\left (\frac{V_i^j[\alpha_0] \cdot V_i^j[\alpha_i] }
{\|V_i^j[\alpha_0]\| \cdot\|V_i^j[\alpha_i]\| +\epsilon} \right)} \end{equation}

For numerical stability, $\epsilon$ is added to the denominator. 
$ang[i]$ is then compared to all previous angles
$\{ang[k] ~ \mid ~  k \leq i \}$.
If the current angle is larger than all previous angles, the current point is visible from
the light source location and thus has no shadow. Otherwise, 
the point (pixel) is shadowed.  
A visualization of the process can be seen in Fig. \ref{fig:viewshed}.

This process can be reduced to a cumulative maximum function as defined in \Cref{eq:cummax}.
To achieve the final shadow estimate in \Cref{eq:shadow_estimate} 
we use a Sigmoid function to keep the final values between 0 and 1. 

\begin{equation}\label{eq:cummax}
s_L[\alpha_i] = \max\big(ang[\alpha_i], s_L[\alpha_{i-1}]\big)
\end{equation}
\begin{equation}\label{eq:shadow_estimate}
s[\alpha_i] = 2 \sigma{\Big(ang[\alpha_i] - s_L[\alpha_i]\Big)}
\end{equation}

The cumulative maximum function was chosen since it is differentiable, 
similar to the well-known ReLU function.
The rest of the process is also differentiable,
and can be used to learn the depth map similar to what is done in inverse rendering methods.

\subsection{Learning Depth from Shadows}
For a given light source $L^j$, a dense estimated shadow map is produced by generating 
the shadow prediction for each pixel. 
The shadow line scan algorithm is applied for many lines in the image, covering the 
entire set of pixels, in order to generate a dense shadow map $\hat{S}^j$.
The ground-truth shadow map can then be used for supervision, to learn the depth.
Once finished, the optimized MLP can be used to generate a dense predicted depth map.



\subsection{Computational Aspects}
In \cite{mildenhall2020nerf} and similar follow-up works, to predict the pixel color, 
an integral must be calculated along a ray of samples.
This requires quadratic computation complexity relative to the image size.
In our work, the calculation takes linear time, since shadow predictions for
all pixels along a shadow line scan can be calculated in a cumulative manner.
The intermediate shadow results for a particular pixel and the maximum angle thus far are
stored and used in the next pixel calculation along the line.
We used the boundary sampling method (R2) \cite{franklin1994higher}, 
which sends LOS rays from the light source to the boundary of image only, instead of 
sending rays to all pixels.
Although this method is considered less accurate than 
the R3 \cite{shapira1990visibility} method which calculates results for all pixels, 
it requiring an order of magnitude fewer calculations.
The R2 method can also be sub-sampled using a coarse-to-fine sampling scheme.
In the early iterations, rather than taking all pixels in the boundary,
one can take every $k_{th}$ 
pixel. Since the light locations vary, we will most likely use all pixels in the depth map 
at least once.


\subsection{Loss Function}
We used a loss composed of the reconstruction loss
and a depth-guided regularization loss. The general loss is: 
\begin{equation}\label{eq:loss}
\mathcal{L}= \frac{1}{N}\sum_{j}\mathcal{L}_{rec}^j + \lambda \mathcal{L}_{d}
\end{equation}
where $\mathcal{L}_{rec}^j $ is the reconstruction loss:

\begin{equation}\label{eq:shadow_loss}
\mathcal{L}_{rec}^j=  \frac{1}{HW} |S^j - \hat{S^j}|.
\end{equation}

$\mathcal{L}_{d}$ is a depth regularization term:
\begin{equation}\label{eq:depth_reg}
\mathcal{L}_{d} = \sum_{ij}\left | \partial_x \hat{d}_{ij} \right | e^{-\left \|
        \partial_x \overline{I}_{ij} \right \|} + \left |
        \partial_y \hat{d}_{ij} \right | e^{-\left \| \partial_y \overline{I}_{ij} \right \|},
\end{equation}
where $\hat{d}_{ij}=\hat{d}(u_i,v_j)$ is the estimated depth at pixel $(u_i, v_j)$,
$\partial_x$ and $\partial_y$
are gradients in the horizontal and vertical directions, and $\overline{I}$ 
is the average color over all
input images. Similar to \cite{DBLP:journals/corr/abs-1903-00112}, 
we assume that the average image edges provide a signal for the object discontinuity. 
This regularization term helps the depth map to converge faster.


\section{Experimental Results}

We compare our method to several shape-from-shading methods\footnote{Shape from \textit{shadow} was previously studied (\Cref{sec:related}). However, all works precede the deep learning era and only target simple objects. We compare to shape from \textit{shading} methods that can be applied to the complex shapes in our datasets.}.
While these methods require illuminated images as input (and we only require binary shadow maps) we find that this comparison illustrates the complementary nature of our method to shape-from-shading. As we will show, results depend on the statistics of the input objects, and for several classes of objects we achieve superior results to shape-from-shading even though we use less data.



\subsection{Implementation details}
We represent the continuous 2D depth map as a function of $\mathbf{u}=(u, v)$.
Similar to  \cite{mildenhall2020nerf}, we approximate this function with
an MLP $F_w(\mathbf{u})=\hat{d}$.
We used six-layer MLP with a latent dimension of 128, 
along with \textit{sine} activation functions, similar to \cite{sitzmann2019siren}. 
The MLP was initialized according to \cite{sitzmann2019siren}. 

We implemented this using PyTorch \cite{NEURIPS2019_9015}. We optimized the neural network
using the Adam optimizer \cite{kingma2017adam} with an initial learning rate of 5~x~$10^{-5}$,
which was decreased every 15 epochs by a factor of 0.9.
We reduce the temperature of the Sigmoid function 3 times during the optimization process, 
using a specific schedule.
We run the optimization until the loss plateaus, which typically takes an hour on
an input of 16 images with spatial resolution of 256x256, using an Intel i9-10900X CPU and an
NVIDIA GeForce RTX 2080 Ti. 
The current implementation uses a single-process and contains many occurrences of native Python
indexing, thus is sub-optimal and can be further optimized.

Photometric-stereo data is not usually accompanied by labeled shadow masks.
Thus, for datasets which have no ground-truth shadows or light directions, 
we implement a model to estimate these.
We use the Blobby and Sculptures datasets from
\cite{ps_fcn} along with our own shadow dataset in order to train the model. 
Details can be found in the supplementary material.

\subsection{Datasets}
We show results on nine synthetic objects: 3 from previous work \cite{kaya2021uncalibrated}
and 6 curated and rendered by us from Sketchfab\footnote{https://sketchfab.com/}. 
The objects we selected span various shapes and sizes, and they also possess one key
property: many 3D features that cast shadows. As we will show, when this property holds, 
our method outperforms shape-from-shading methods and thus complements them.
Our objects were each rendered with 16 different illumination conditions. 
Scenes are illuminated by a single point-light source.

Since shadow maps are not available in the dataset of
Kaya~et~al.~\cite{kaya2021uncalibrated}, we have our previously mentioned shadow-extraction
model to estimate these. For our rendered objects, 
we use the ground-truth shadow maps as inputs.
Qualitative results on real objects (without ground-truth) are available in the
supplementary material.

\begin{figure}[t] \centering
    \makebox[0.15\textwidth]{\scriptsize Relief}
    \makebox[0.15\textwidth]{\scriptsize Sculptures}
    \makebox[0.15\textwidth]{\scriptsize Cactus}
    \makebox[0.15\textwidth]{\scriptsize Surface}
    \makebox[0.15\textwidth]{\scriptsize Bread~~~}
    \makebox[0.15\textwidth]{\scriptsize Rose~~~~}
    \begin{tabular}{lc}
        \rotatebox{90}{\scalebox{.7} {~Ground Truth}}
        \includegraphics[width=0.9\textwidth]{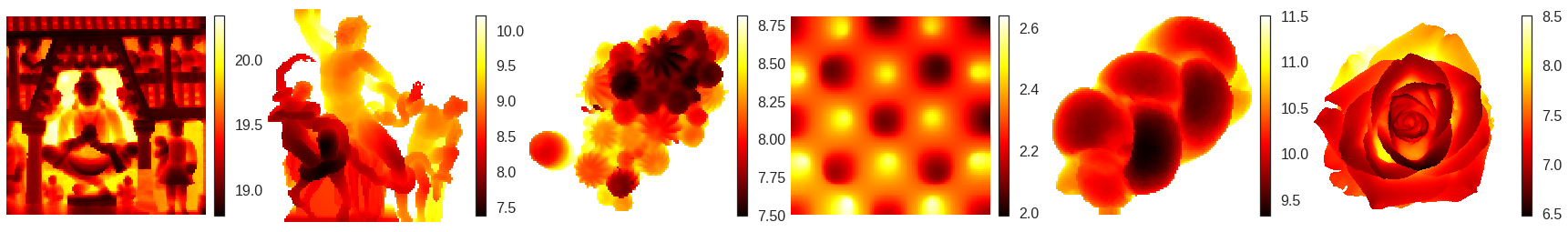}
    \end{tabular}
    
    \begin{tabular}{lc}
        \rotatebox{90}{\scalebox{.7} {~ Santo et al.}} 
        \includegraphics[width=0.9\textwidth]{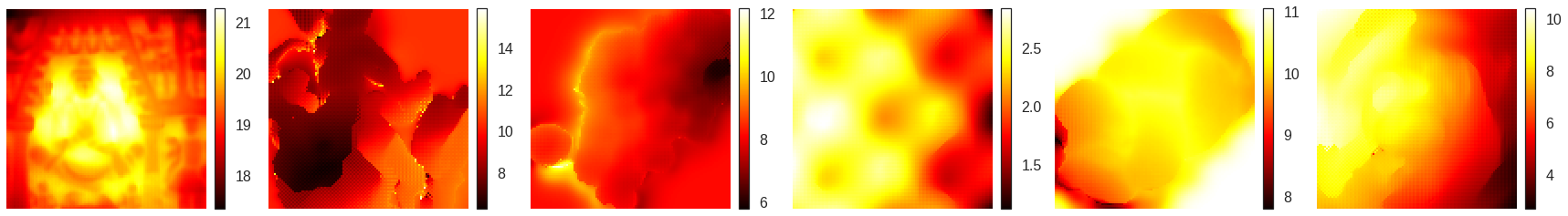}
    \end{tabular}
    
    \begin{tabular}{lc}
        \rotatebox{90}{\scalebox{.7} {~~  Peng et al.}}
        \includegraphics[width=0.9\textwidth]{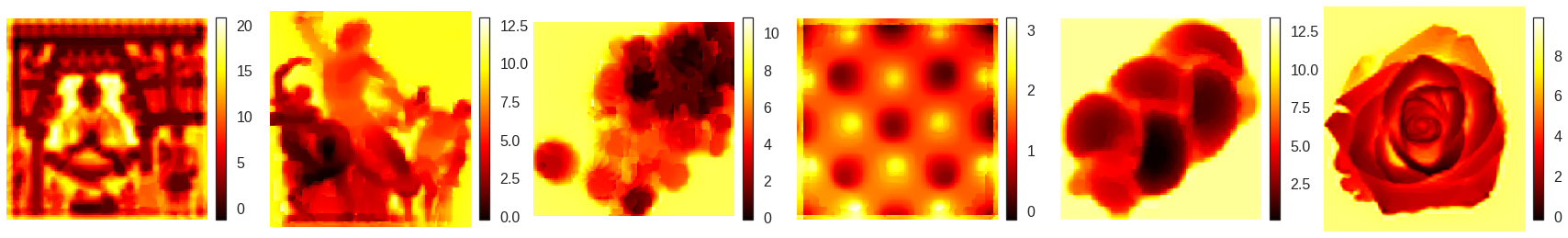}
    \end{tabular}
    
    \begin{tabular}{lc}
        \rotatebox{90}{\scalebox{.7} {~~  Ours}}
        \includegraphics[width=0.9\textwidth]{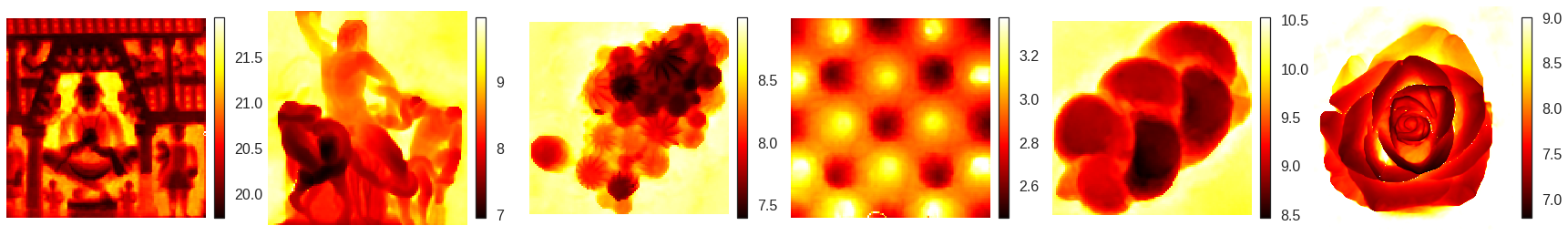}
    \end{tabular}
    
    \caption{Comparison of depth maps. Each row refers to a specific method. } 
    \label{fig:visual_results_depth}
\end{figure}

\begin{figure}[t] 
    \centering
    
    \makebox[0.15\textwidth]{\scriptsize Relief}
    \makebox[0.15\textwidth]{\scriptsize Sculptures}
    \makebox[0.15\textwidth]{\scriptsize ~~Cactus}
    \makebox[0.15\textwidth]{\scriptsize Surface}
    \makebox[0.15\textwidth]{\scriptsize Bread}
    \makebox[0.15\textwidth]{\scriptsize Rose}
    \\
    \begin{tabular}{lc}
        \rotatebox{90}{\scalebox{.7} {~~ Ground Truth}}
        \includegraphics[width=0.9\textwidth]{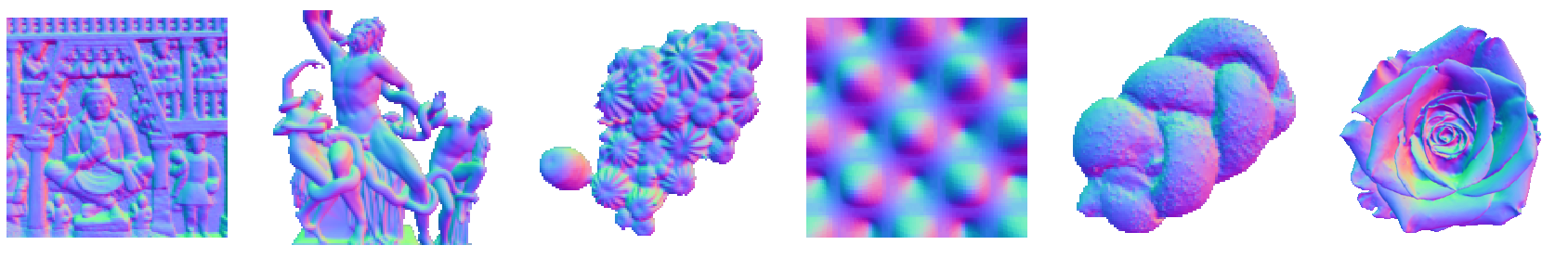}
    \end{tabular}
    \\
    \begin{tabular}{lc}
        \rotatebox{90}{\scalebox{.7} {~~ Santo et al.}}
        \includegraphics[width=0.9\textwidth]{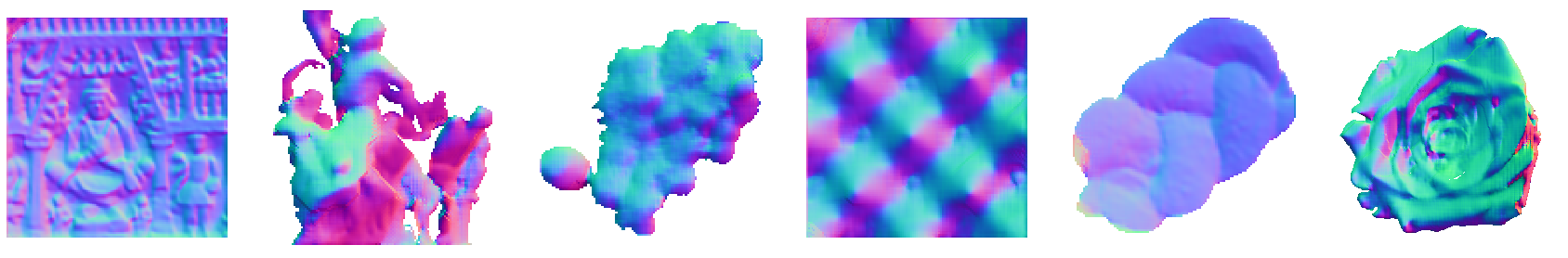}
    \end{tabular}
    \\
    \begin{tabular}{lc}
        \rotatebox{90}{\scalebox{.7} {~~ Chen et al.}}
         \includegraphics[width=0.9\textwidth]{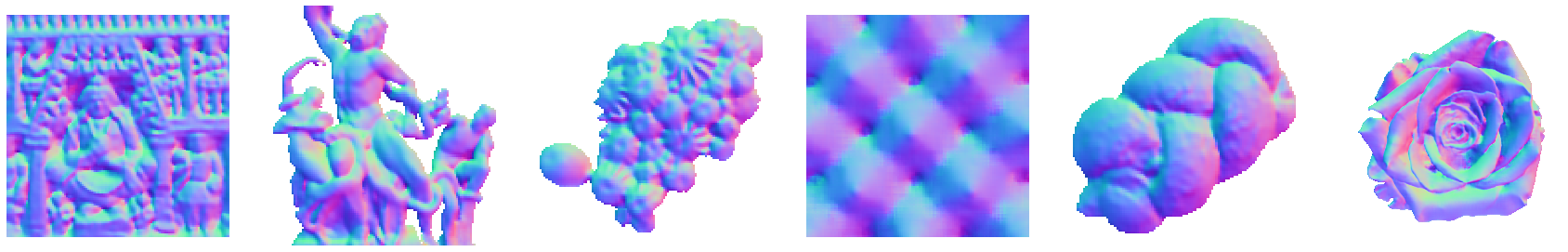}
    \end{tabular}
    \\
    \begin{tabular}{lc}
        \rotatebox{90}{\scalebox{.7} {~~ Ours}}
        \includegraphics[width=0.9\textwidth]{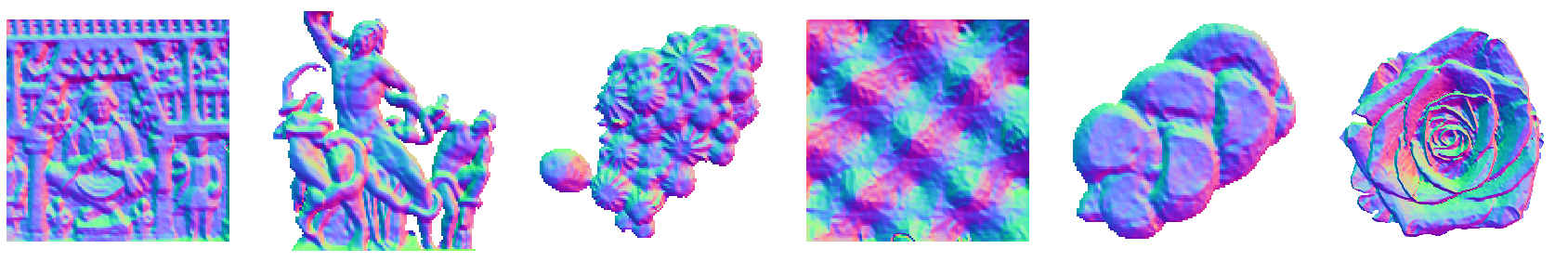}
    \end{tabular}
    \\
    \caption{Comparison of surface normals estimated by each of the compared methods.
    Each row represents a different method.} 
    \label{fig:visual_results_normals}
\end{figure}

\subsection{Comparisons}
We compare to the state-of-the-art methods of Chen~et~al.~\cite{chen2019selfcalibrating}, Santo~et~al.~\cite{DeepNearFieldPS}, and Peng~et~al.~\cite{peng2017iccvw},
that use photometric stereo inputs, with the latter requiring a down-sampled depth map as an additional input.
We compare both depth maps \cite{peng2017iccvw,DeepNearFieldPS} and normals
\cite{chen2019selfcalibrating,DeepNearFieldPS} (according to the generated
outputs of each method).
%
%

\Cref{fig:visual_results_depth} shows a qualitative evaluation of depth map result quality.
The method of Santo~et~al. produces a low frequency estimation of depth, but does not
contain any high frequency details. The method of Peng~et~al. produces results which are
closer to the ground truth, but still lacking in sharpness (e.g., in \textit{Relief} and
\textit{Cactus}). Our method is overall closest to the ground truth depth. Quantitatively,
we measure normalized mean depth error (nMZE)
\footnote{we normalize each depth map by its own std and bias, in order to make the
comparisons scale and bias invariant} in \Cref{tab:Our_dataset}. 
We achieve lower nMZE on 5 out of the 6 objects, with an overall average nMZE which is 1.7
times \cite{peng2017iccvw} and 4.9 times \cite{DeepNearFieldPS} lower than the alternative.


\Cref{fig:visual_results_normals} shows a qualitative evaluation of surface normal result quality.
The method of Santo~et~al. produces overly smooth normals, with some examples relatively close to the ground truth (e.g., \textit{Relief}) but others farther away (e.g., \textit{Rose}). The method of Chen~et~al. is comparable to ours, with generally good results that resemble the ground truth normals, but lack sharp transitions. 
Quantitatively, we measure mean angle error (MAE in degrees) in \Cref{tab:Our_dataset}. 
Our results are comparable to (albeit slightly better than) those of Chen~et~al., 
and 1.6 times better than those of Santo~et~al.

\begin{table}
    \centering
    \caption{Quantitative evaluation. The top three rows are depth map results, 
    and the bottom rows are surface normal results. All methods but Peng~et~al. were given
    ground-truth light location, and Peng~et~al. was given a down-scaled depth map.}
    \begin{tabular}{l || c | c c c c c c || c}
    \toprule
    Method  & Metric & Cactus  &  Rose & Bread & Sculptures & Surface   & Relief     & Avg\\ 
    \midrule
    Santo~et~al.~\cite{DeepNearFieldPS} & nMZE & 0.96           & 1.16           & 0.77           & 0.81           & 0.75           & 0.75           & 0.87            \\ 
    Peng~et~al.~\cite{peng2017iccvw}    & nMZE & 0.43           & \textbf{0.05}  & 0.40           & 0.20           & 0.33           & 0.42           & 0.31            \\ 
    Chen~et~al.~\cite{chen2019selfcalibrating}& nMZE & N/A            & N/A            & N/A            & N/A            & N/A            & N/A            & N/A             \\
    Ours                                & nMZE & \textbf{0.33}  & 0.11           & \textbf{0.16}  & \textbf{0.19}  & \textbf{0.10}  & \textbf{0.18}  & \textbf{0.18}   \\  
    \midrule
    Santo~et~al.~\cite{DeepNearFieldPS} & MAE  & 32.79          & 50.23          & 33.95          &  54.49         & 22.21          & 21.93          & 35.93           \\ 
    Peng~et~al.~\cite{peng2017iccvw}    & MAE  & N/A            & N/A            & N/A            & N/A            & N/A            & N/A            & N/A             \\
    Chen~et~al.~\cite{chen2019selfcalibrating} & MAE  & 24.61          & \textbf{25.12} & \textbf{18.31} & \textbf{26.43} & 18.91          & 25.46          & 23.14           \\
    Ours                                & MAE  & \textbf{22.60} & 26.27          & 20.43          &  27.50         & \textbf{17.16} & \textbf{21.80} & \textbf{ 22.63} \\  
    \bottomrule
    \end{tabular}

    \label{tab:Our_dataset}
\end{table}
\begin{table}[t]
    \centering
    \caption{Comparison of surface normal estimation on the dome dataset in
    \cite{kaya2021uncalibrated} using calibrated methods (known light locations). 
    Compared results were used from the original paper.}
    \begin{tabular}{l || c | c c c }
    \toprule
 Method            & Metric   & Vase        & Golf Ball & Face \\ 
 \midrule
 Chen~et~al.~\cite{ps_fcn} & MAE & 27.11          & 15.99          & 16.17.81          \\
 Kaya~et~al.~\cite{kaya2021uncalibrated} & MAE & 16.40   & \textbf{14.23} & \textbf{14.24} \\  
 Ours                                    & MAE & \textbf{11.24} & 16.51          & 18.70          \\
 \bottomrule
    \end{tabular}
    \label{tab:dome_dataset}
\end{table}
\begin{table}[t]
    \centering
    \caption{Average shadow pixels in image analysis. We measure MAE of estimated surface normals for increasing amount of average shadows. These are calculated by taking the average amount of shadowed pixels in each rendered object.}
    \begin{tabular}{l || c c c c c || c}
    \toprule
 Avg. Shadow &   A(48\%)  & B(60\%)   & C(66\%) & D(69\%)   & E(60\%)* & Avg  \\ 
 \midrule
 Chen~et~al.~\cite{chen2019selfcalibrating}& 18.63          & 18.71          & 20.43          & 21.44          & 22.99          & 20.44          \\
 Santo~et~al.~\cite{DeepNearFieldPS} & \textbf{16.18} & \textbf{17.25} & \textbf{17.19} & 18.25          & 42.70          &  22.31         \\
 Ours         & 20.43          & 18.53          & 17.95          & \textbf{17.53} & \textbf{18.68} & \textbf{18.62} \\
 \bottomrule
    \end{tabular}
    \label{tab:diff_light_dists}
\end{table}

Note that \cite{kaya2021uncalibrated} refers to a \textit{rose} object as an object 
with `complex geometry and high amount of surface discontinuity', 
where their method fails due to their assumption of a continuous surface and due to a
high amount of cast shadows.
We render a similar object and show that our method does not require such assumptions.
Visual results can be seen in \Cref{fig:banner}.

Besides our own generated dataset, we compare our results to three different shape-from-shading results reported in \cite{kaya2021uncalibrated}
\footnote{The code was not available at the time of writing this paper, so the reported results in \cite{kaya2021uncalibrated} were used as-is.}
which can be seen in \Cref{tab:dome_dataset}. 
Since the \textit{vase} object is concave, it has a strong cast shadow
affect, which is a strong learning signal. We can also see that the shape-from-shading 
methods fail to deal with this. 
The \textit{golf} and \textit{face} objects have sparse cast shadows, 
which is likely the reason why our method does not perform well there.

\subsection{Analysis \& Ablation}
\textbf{Performance on various shadow amounts}
To further test our method, we generated the \textit{surface} object using 16
lights placed at different incidence angles, thus creating inputs with
increasing amount of shadows. We rendered this by placing the object at a constant location,
and increasing the distance of the lights from the object at each different scene.
We observe in \Cref{tab:diff_light_dists} that DeepShadow's accuracy improves 
as the amount of shadow present in the image increases, 
as opposed to the other tested methods, in which the accuracy degrades as the amount of
shadowed pixels increases.
Note that objects \textit{A--D} were rendered with lights at a constant distance, and object
\textit{E} was rendered with lights at varying distances. We observe that
\cite{DeepNearFieldPS} does not handle this scenario well.



\begin{figure}[t] \centering
    \makebox[0.25\textwidth]{\scriptsize Cactus}
    \makebox[0.3\textwidth]{\scriptsize Surface}
    \makebox[0.2\textwidth]{\scriptsize Bread}
    \makebox[0.1\textwidth]{\scriptsize  }
    \\
    \rotatebox{90}{\tiny~ Depth map error}
    \includegraphics[width=0.9\textwidth]{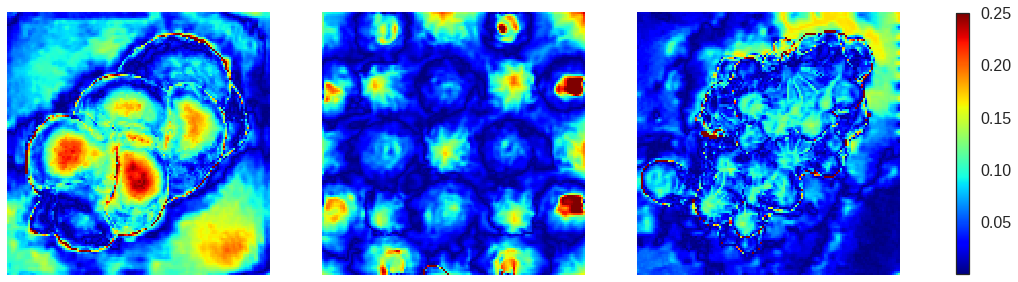}
    \\
    \rotatebox{90}{\tiny~ Sum of shadowed pixels}
    \includegraphics[width=0.9\textwidth]{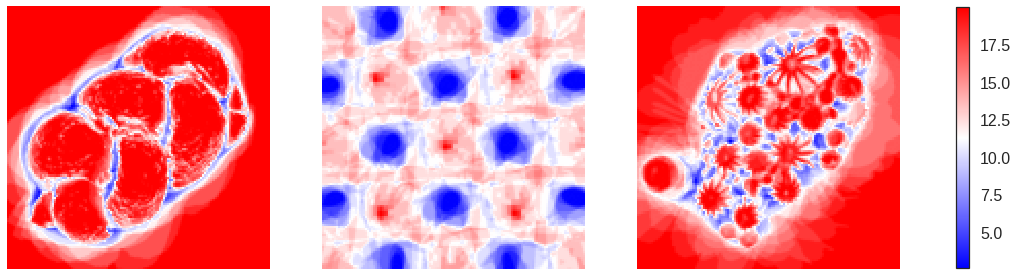}
    \\
    \caption{Comparison of Depth Estimation Error (top) vs. Number of Shadowed Pixels (bottom).
    In the bottom row, we can observe pixels that are mostly shadowed 
    (blue) or mostly illuminated (red) in all renderings. 
    These pixels have higher depth estimation errors (top row).
    Pixels that are balanced, i.e., 
    shadowed and illuminated in roughly equal proportions (white, bottom row)
    have a smaller depth estimation error.} 
    \label{fig:num_shadow_pixels_ablation}
\end{figure}

\textbf{Shadowed and illuminated pixels.} We analyze the effects of the number of 
shadowed and illuminated pixels on the final depth estimation.
In \Cref{fig:num_shadow_pixels_ablation} we compare the error map (top) to 
the number of illuminated pixels across all lighting conditions (bottom). 
Our method works best if this sum is balanced, i.e., 
if pixels are illuminated in some samples and shadowed in other samples.
If pixels are always shadowed or always illuminated, reconstruction is ill-posed and our method does not perform well.



\subsection{Failure Cases}
Our algorithm is based on the line of sight algorithm, 
thus it relies on shadow gradients to reconstruct the
underlying depth map. 
In objects such as the \textit{face} in \Cref{tab:dome_dataset}, the algorithm fails in flat
areas, e.g., as the forehead, since the latter is relatively smooth and convex - 
and thus has no shadow from which to learn from.
This is also true for the \textit{bread object} in \Cref{tab:Our_dataset},
which has large areas that are flat, as can be seen in \Cref{fig:num_shadow_pixels_ablation}.
In contrast, the \textit{vase} in \Cref{tab:dome_dataset} is relatively smooth yet also
concave, and has many cast shadows, which enables our method to succeed. 

\section{Conclusions}

In this work we proposed DeepShadow, a deep-learning based method for recovering 
shape from shadows.
We show that in various scenarios, the shadow maps serve as good learning signals that enable 
the underlying depth and surface normals to be recovered.
Experiments have shown that this method achieves results equal to or better than those generated by the various shape-from-shading
algorithms, for severely shadowed objects, while using fewer or no data --- 
since we are not using any training data besides the shadow maps.
An additional benefit of our method is that we do not require knowing the light intensity, 
since the shadow maps are of binary values. Even though previous works have shown this can be 
estimated, the estimation may be erroneous, which affects the final result.

Our method fails on convex objects or objects with sparse shadows. 
In future work, this method should be combined with shape-from-shading methods in order for 
both methods to benefit from each other.
Another possible research direction is producing super-resolution depth maps using the shadow clues, which 
would combine our work and \cite{peng2017iccvw}. Since we are using implicit representations, this may work well.

\subsubsection{Acknowledgements}

This work was supported by the Israeli Ministry of Science and Technology under The National
Foundation for 
Applied Science (MIA), and by the Israel Science Foundation (grant No. 1574/21).

\printbibliography   

\appendix
\newcommand{\hbAppendixPrefix}{A}
\renewcommand{\thefigure}{\hbAppendixPrefix\arabic{figure}}
\setcounter{figure}{0}
\renewcommand{\thetable}{\hbAppendixPrefix\arabic{table}} 
\setcounter{table}{0}
\renewcommand{\theequation}{\hbAppendixPrefix\arabic{equation}} 
\setcounter{equation}{0}

\def\ShowNotes{}

\section{Shadow and light extraction network}
\label{sec:shadow-net}

Since shadow maps are not always available, we train a network to estimate them 
from input photometric stereo images. This network runs as a pre-processing step on the 
photometric stereo data, to produce shadow maps that DeepShadow can use as inputs.
Our model also estimates the light direction, since this is also not always available.
We use both a publicly available photometric stereo dataset, as well as our own 
renders --- which are needed
since there is no public dataset that has photometric stereo shadow ground-truth.

Please note that our goal is estimating depth from shadow maps. As we have shown, in certain cases DeepShadow with shadow maps as inputs may result in better shape estimation than
shape-from-shading techniques. 
We trained the shadow and light extraction model solely to be able to use our method on datasets which do not
have ground-truth shadow maps or light directions, so that we are able to test our method on 
more types of objects and scenes.
Using the light extraction model, we estimate the direction of lights (located at infinity), 
and then convert these to point-lights by projecting onto the unit sphere and multiplying by a constant, empirically set to twice the distance between camera and object.

\subsection{Network Architecture}
\label{subsec:arch}

\begin{figure}[t]
\centering
\includegraphics[width=0.9\textwidth]{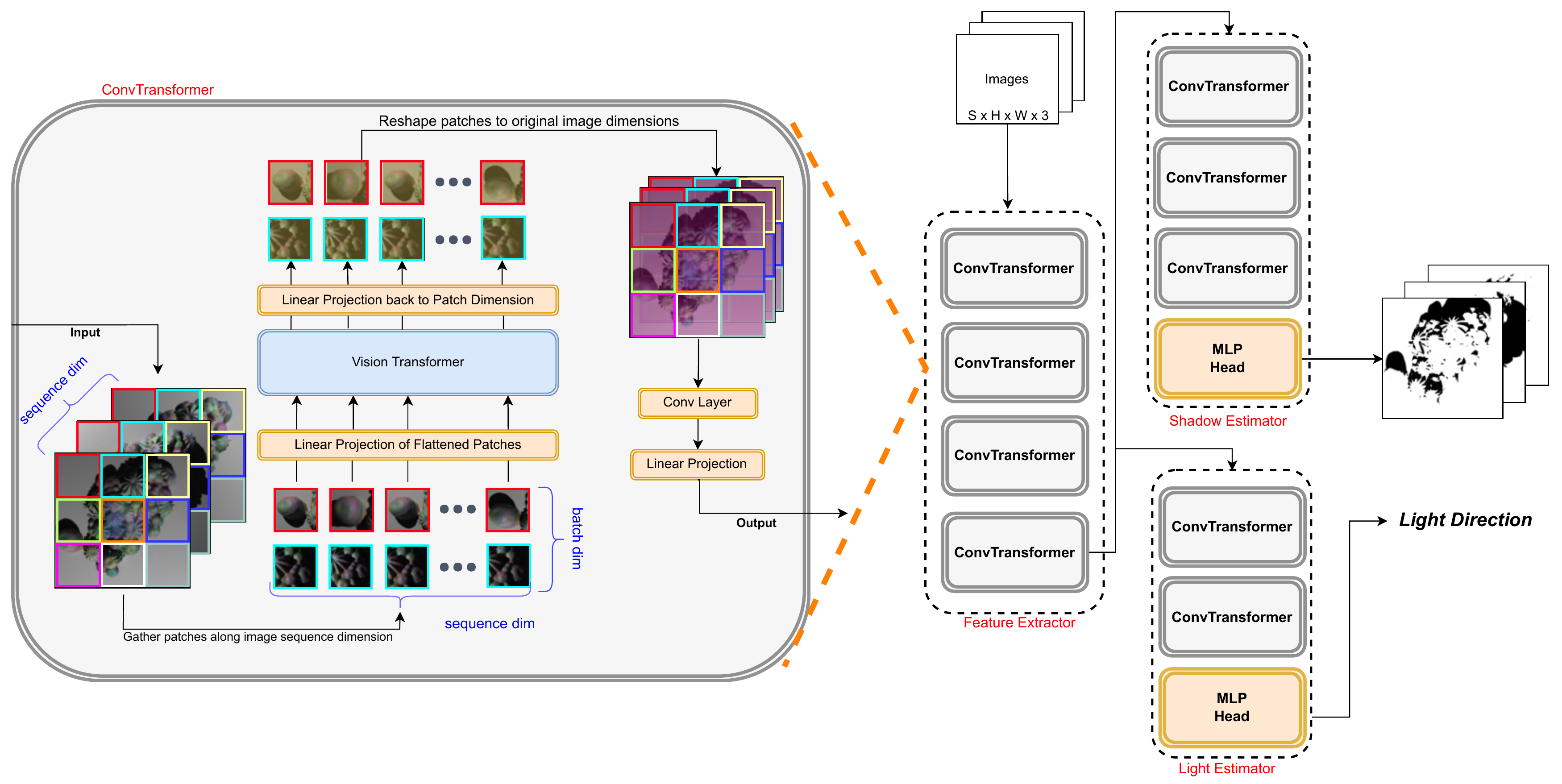}
\caption[.]
{Shadow Estimation Model. The model receives as inputs a sequence of photometric stereo 
images and outputs the estimated light directions and shadow maps. 
Each image is split to patches, and patches are gathered by the sequence index. Each such
sequence is fed separately to the ConvTransformer.
The model is composed of a four layer ConvTransformer which outputs intermediate features. 
These features are used to generate the final outputs, using 2 smaller ConvTransformers. 
The ConvTransformer can be viewed in detail in the bottom of the figure.}
\label{fig:shadow_transformer}
\end{figure}

The shadow estimation model is illustrated in \Cref{fig:shadow_transformer}.
It is used for estimating light directions and shadow maps given
photometric stereo image inputs. Although the model also outputs normal maps, these
are only used during the training and are discarded during the inference.
The input images have dimensions of $S \times C \times W \times H$, 
$S$ being the number of input images (sequence
dimension), $C$ is the image color channel, and $H \times W$ is the spatial image size.

The complete model is composed of four hybrid Transformer-Convolution layers (ConvTransformers):
the first block is a  ConvTransformer for extracting features from the input images,
and the second and third blocks are two ConvTransformers for estimating shadows and
for estimating light directions.
The last custom block is used for estimating the normals from the features (not illustrated in 
\Cref{fig:shadow_transformer}).

The feature-extraction ConvTransformer splits each input image into an 8x8 patch,
as applied in Vision Transformer (ViT) \cite{VIT}. 
In contrast to the regular ViT, we group all
patches along the \textit{sequence} dimension, since the main target of the model is predicting
per-pixel output for each patch. 
Since predicting shadows from photometric stereo images is essentially a threshold-based
problem, we choose to use an attention-based model and compare all spatially co-located
patches, instead of comparing all patches in a single image.
These patches are then used as a sequence of inputs to the transformer.
Each sequence is passed separately through a ConvTransformer 
(sequences can be batched together) to produce a sequence of intermediate features 
extracted from the relevant patches.
Once all the patches have passed through the transformer, 
they are reshaped back to the
original image dimensions, and then passed through a convolution layer to output the
features. 
The feature-extraction ConvTransformer is composed of 4 Transformer-Convolution
pairs, with 16 attention heads and latent MLP dimension of 1024.

The light direction block uses 2 ConvTransformer blocks, with a 128 dim MLP and 
4 attention heads. 
The shadow estimation block uses 3 ConvTransformer blocks with 6 attention heads
and an MLP dimension of 128. 
It utilizes a Sigmoid function for outputting values between 0 and 1.
We also estimate the surface normals from the features, using a linear projection layer and two
convolution layers. This is done in order to be able to learn from datasets which have
photometric stereo data and ground-truth normals.

\subsection{Training Details}
We use the Blobby and Sculptures datasets \cite{ps_fcn}, which contain photometric
stereo images, light directions and surface normals.
We also render a shadow dataset composed of 10 objects downloaded from 
Sketchfab\footnote{https://sketchfab.com/}. 
Each object was rendered in 32 different view
angles and with 32 light angles for each view. We use Blender \cite{Blender} to render our datasets.
We use all three datasets during training in a ratio of 1:1:10, i.e.,
for every 10 iterations over the shadow dataset, we iterate once over the Blobby and Sculptures datasets.

During training, we randomly crop each input images to $64 \times 64$. We use random noise and color jitter augmentations,
as well as randomize the sequence length between 16 and 32 inputs, in order to make the transformer agnostic to
the number of input images.

We use the following loss function: 
\begin{equation}\label{eq:shadow_transformer_loss}
\mathcal{L} =\frac{1}{M}\sum_{m}\left( \mathcal{L}_N^m  + \mathcal{L}_S^m +  \mathcal{L}_L^m \right)
\end{equation}
where $\mathcal{L}_N^m$ is the normal loss, $\mathcal{L}_S^m$ is the shadow reconstruction loss and 
$\mathcal{L}_L^m$ is the light direction loss.
The sum is performed over all $m\in[0,M]$ sets of photometric images in the dataset. 
Each such set has $k\in[0,K]$ 
images of size $H\times W $ along with the associated light directions $\ell_k$
and a ground-truth normal map $N$. Our rendered dataset
also has ground-truth shadow maps $S_k$.
The complete loss combines between the loss of the ground-truth normals $N$
and the predicted normals $\hat{N}$, 
\begin{equation}\label{eq:normal_loss}
\mathcal{L}_N^m =\frac{1}{HW} (1-N^m\cdot\hat{N}^m),
\end{equation}
the L1 loss of the ground-truth and predicted shadow maps
\begin{equation}\label{eq:shadow_loss_suppl}
\mathcal{L}_S^m =\frac{1}{KHW}\sum_{k}|S^m_k- \hat{S}^m_k|,
\end{equation}
and the cosine embedding loss between the ground-truth light direction
$\ell^m_k$ and the estimated direction $\hat{\ell}^m_k$
\begin{equation}\label{eq:light_loss}
\mathcal{L}_L^m =\frac{1}{K}\sum_{k}(1 - cos(\ell^m_k,\hat{\ell}^m_k)).
\end{equation}
We omit the supervision on the lights when using our dataset, and omit the shadow supervision
when using Blobby and Scupltures datasets. 
We train using the Adam optimizer \cite{kingma2017adam} for 1000 epochs with an initial learning
rate of $1\times10^{-4}$ which is decreased by a factor of 0.8 every 15 epochs.

\begin{figure}[t] \centering
    \begin{tabular}{lc}
        \rotatebox{90}{\scalebox{.7} {\hspace{0.1cm} Input Images}}
        \includegraphics[width=0.95\textwidth]{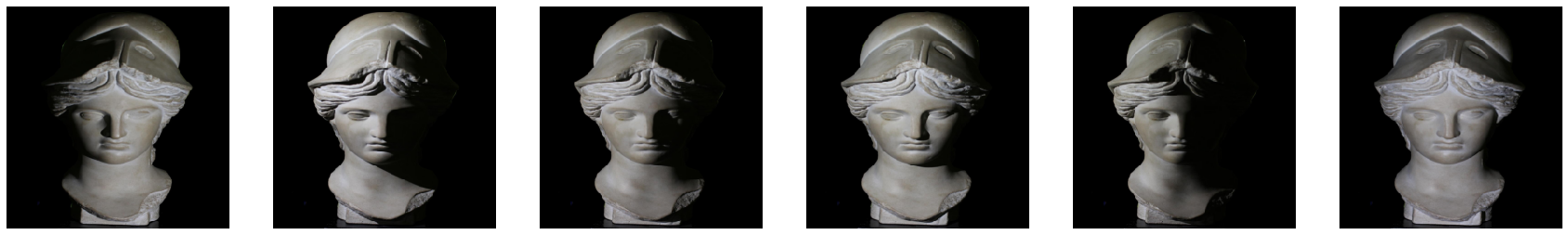}
    \end{tabular}
    
    \begin{tabular}{lc}
        \rotatebox{90}{\scalebox{.7} {\hspace{0.1cm} Est. Shadows}}
        \includegraphics[width=0.95\textwidth]{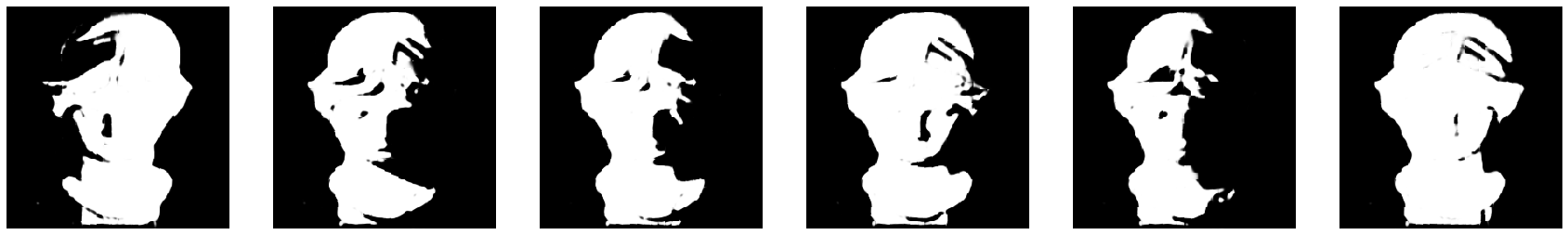}
    \end{tabular}
    
    \caption{Sculpture head object shadow estimation results.}
    \label{fig:pesel_results}
\end{figure}

The results of the shadow estimation can be seen in \Cref{fig:pesel_results}.

\section{Additional Results}
In this section, we present more results of the DeepShadow method.
We also examine the performance of the shadow estimation network on data which lacks
ground-truth shadows.

\subsection{Normal map errors}
\begin{figure}[t]
    \setlength\tabcolsep{1.5pt}
    \newlength{\mycolw}
    \setlength{\mycolw}{0.14\columnwidth}
    \begin{tabular}{cccccccc}
    
        & 
        \scriptsize{Relief} & 
        \scriptsize{Sculptures} & 
        \scriptsize{Cactus} & 
        \scriptsize{Surface} & 
        \scriptsize{Bread} & 
        \scriptsize{Rose} & 
        \\
    
        \raisebox{0.1\normalbaselineskip}[0pt][0pt]{\rot{\scriptsize{\hphantom{ }Santo et al.}}} &
    	\includegraphics[width=\mycolw,frame]{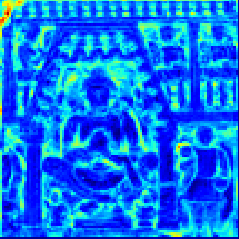} &
    	\includegraphics[width=\mycolw,frame]{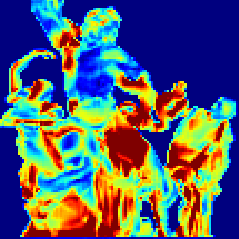} &
    	\includegraphics[width=\mycolw,frame]{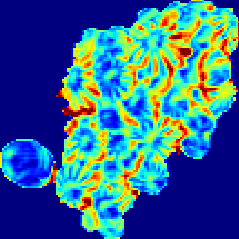} &
    	\includegraphics[width=\mycolw,frame]{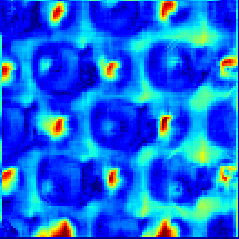} &
    	\includegraphics[width=\mycolw,frame]{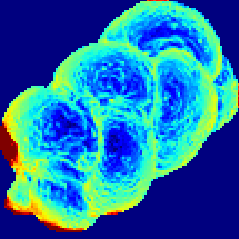} &
    	\includegraphics[width=\mycolw,frame]{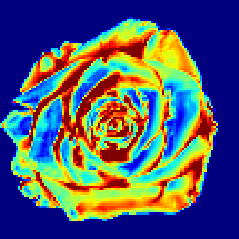} &
    	\raisebox{3.\normalbaselineskip}[0pt][0pt]{\multirow[c]{3}{0em}{\includegraphics[width=0.08\columnwidth]{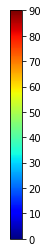}}}
        \\
    	
    	\raisebox{0.1\normalbaselineskip}[0pt][0pt]{\rot{\scriptsize{\hphantom{ }Chen et al.}}} &
    	\includegraphics[width=\mycolw,frame]{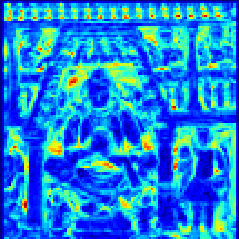} &
    	\includegraphics[width=\mycolw,frame]{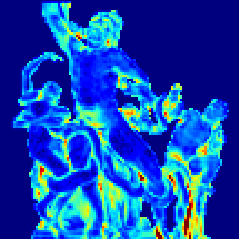} &
    	\includegraphics[width=\mycolw,frame]{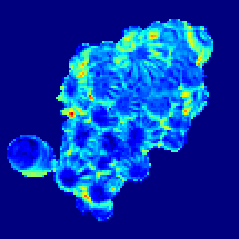} &
    	\includegraphics[width=\mycolw,frame]{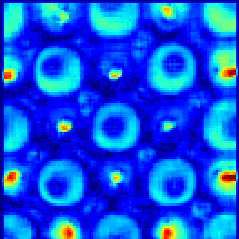} &
    	\includegraphics[width=\mycolw,frame]{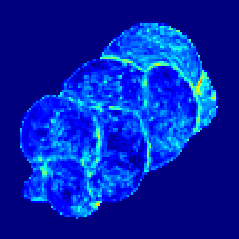} &
    	\includegraphics[width=\mycolw,frame]{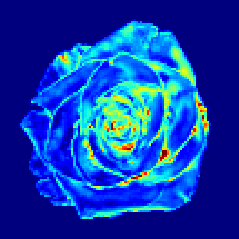} &
        \\
        
        \raisebox{0.8\normalbaselineskip}[0pt][0pt]{\rot{\scriptsize{\hphantom{A.}Ours}}} &
    	\includegraphics[width=\mycolw,frame]{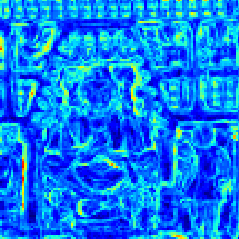} &
    	\includegraphics[width=\mycolw,frame]{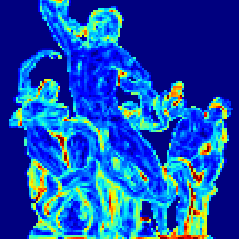} &
    	\includegraphics[width=\mycolw,frame]{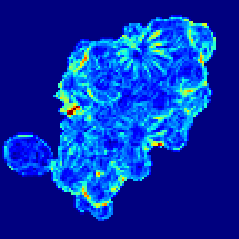} &
    	\includegraphics[width=\mycolw,frame]{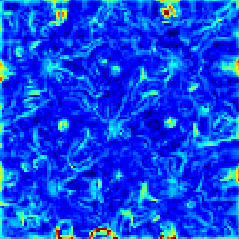} &
    	\includegraphics[width=\mycolw,frame]{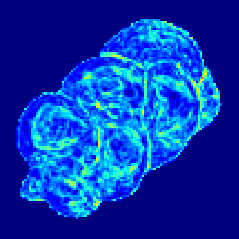} &
    	\includegraphics[width=\mycolw,frame]{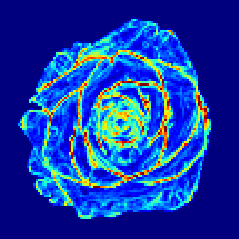} &
        \\

    \end{tabular}


    \caption{Normal error maps comparing our method to \cite{chen2019selfcalibrating} and \cite{DeepNearFieldPS}. }
    \label{fig:visual_results_normals_errors}
\end{figure}

\Cref{fig:visual_results_normals_errors} shows the previously shown objects' normal map errors. 
Our method produces less normal errors on the \textit{Relief}, \textit{Cactus}, \textit{Surface} and \textit{Rose} objects compared to other methods.

\subsection{Specular and diffuse objects}

\begin{figure}[t]
    \centering
    \setlength\tabcolsep{1.5pt}
    \setlength{\mycolw}{0.16\columnwidth}
    \begin{tabular}{cccccc}

    	\scriptsize{Diffuse Obj.} &
        \scriptsize{Specular Obj.} &
        \scriptsize{SDPS Diff.} & 
        \scriptsize{SDPS} Spec. & 
        \scriptsize{Ours Diff.} & 
        \scriptsize{Our Spec.}   
        \\
        
    	\includegraphics[width=\mycolw,frame]{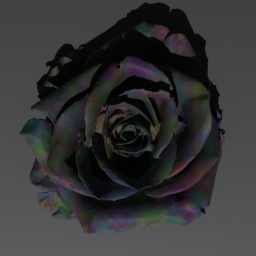} &
    	\includegraphics[width=\mycolw,frame]{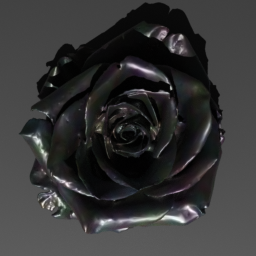} &
    	\includegraphics[width=\mycolw,frame]{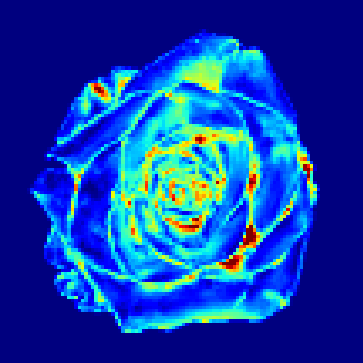} &
    	\includegraphics[width=\mycolw,frame]{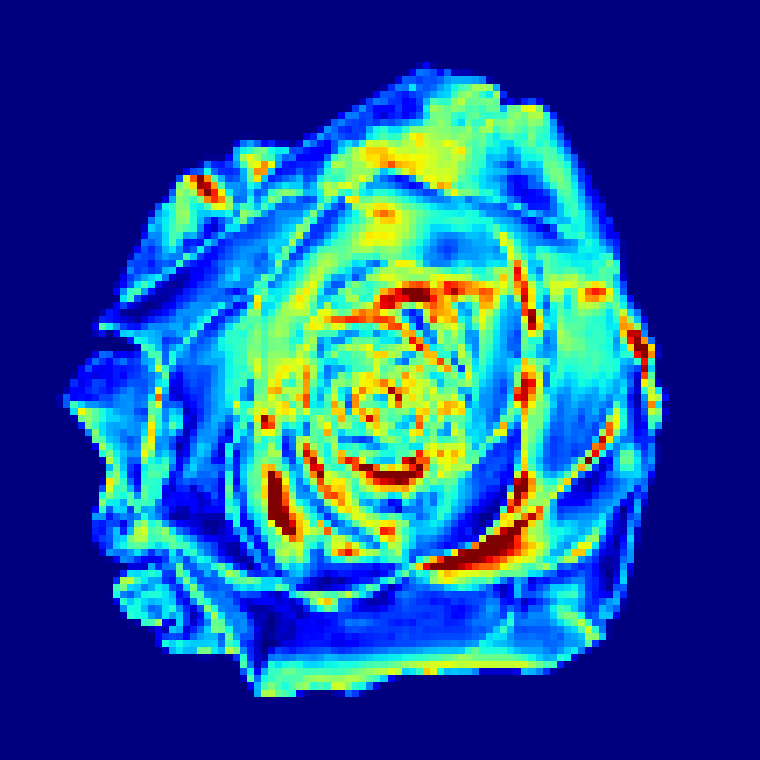} &
    	\includegraphics[width=\mycolw,frame]{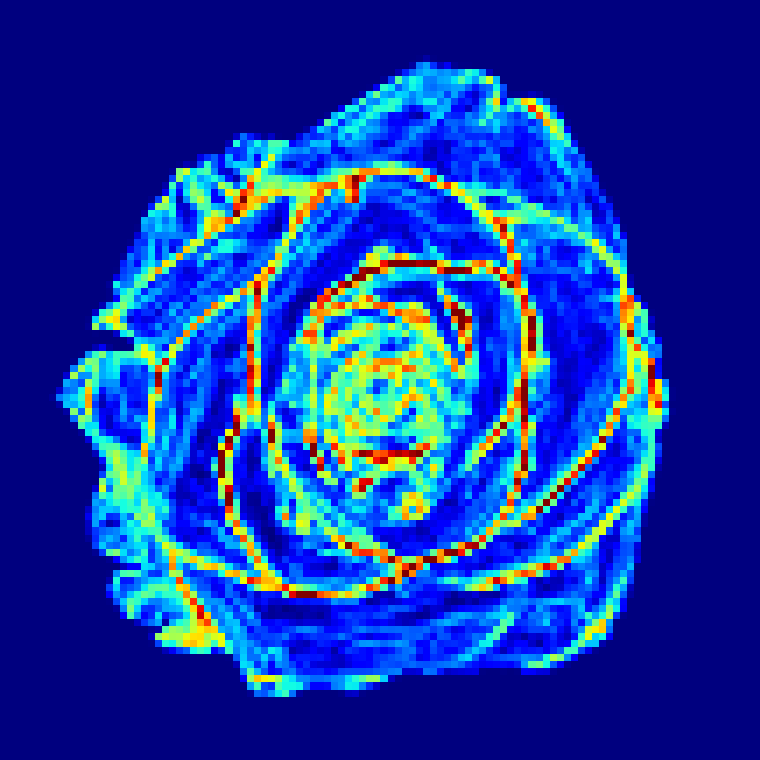} &
    	\includegraphics[width=\mycolw,frame]{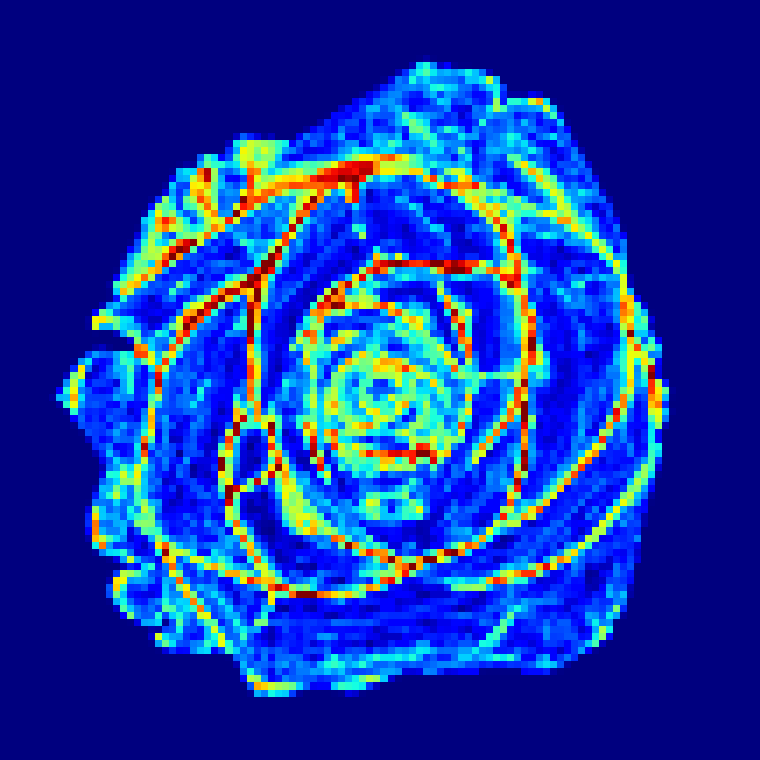} 
        \\

    \end{tabular}

	\vspace{-0.1in}
    \caption{Effect of specularities on angle error of output normals.} 
    \label{fig:visual_results_specular_diffuse}
    \vspace{-0.5cm}
\end{figure}

We test our method's resiliency to specular inputs and compare to that of a shape-from-shading method.
We rendered a modified version of our \textit{rose} object, with two different materials;
highly specular and diffuse (\Cref{fig:visual_results_specular_diffuse}). 

We evaluated our algorithm (including the shadow extraction model) on these objects, and compared the 
surface normal results to SDPS~\cite{chen2019selfcalibrating}.
The diffuse object produces 25.02 MAE and the metallic object produces 34.35 MAE using SDPS, 
while our method achieves 26.50 MAE and 27.76 MAE, respectively. 
With specularity added, we can see a big drop in accuracy using SDPS, while our method achieves a smaller drop.

\subsection{Shadow maps reconstruction error}
\begin{figure}[t] \centering
    \begin{tabular}{lc}
        \rotatebox{90}{\scalebox{.65} {\hspace{0.1cm} GT Shadows}}
        \adjincludegraphics[width=0.95\textwidth,Clip={0\width} {0.5\height} 
        {0\width} {0\height}]{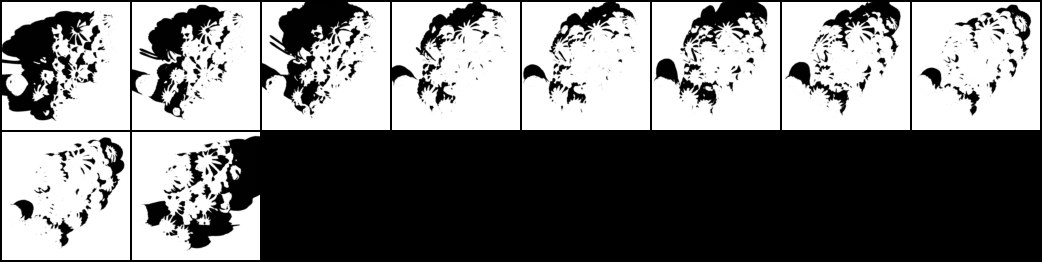}
    \end{tabular}
    
    \begin{tabular}{lc}
        \rotatebox{90}{\scalebox{.65} {\hspace{0.1cm} Est. Shadows}}
        \adjincludegraphics[width=0.95\textwidth,Clip={0\width} {0.5\height} 
        {0\width} {0\height}]{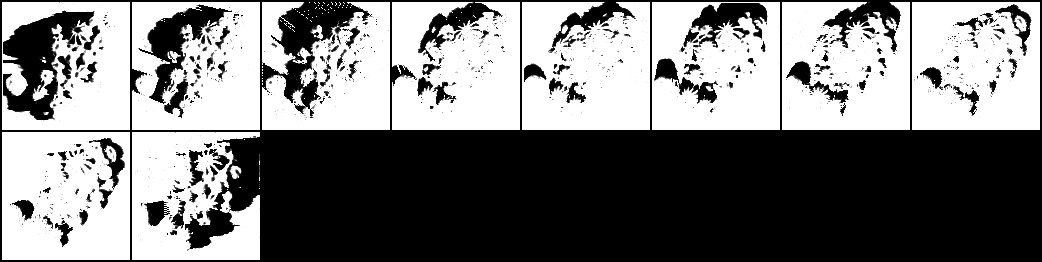}
    \end{tabular}
    
    \begin{tabular}{lc}
        \rotatebox{90}{\scalebox{.65} {\hspace{0.1cm} Shadow Error}}
        \adjincludegraphics[width=0.95\textwidth,Clip={0\width} {0.5\height} 
        {0\width} {0\height}]{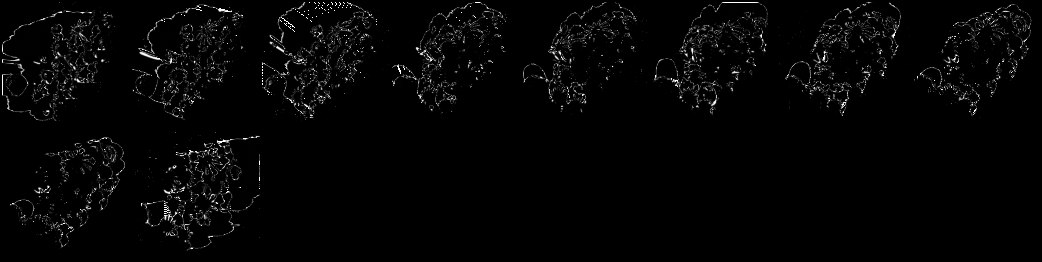}
    \end{tabular}
    
    \begin{tabular}{lc}
        \rotatebox{90}{\scalebox{.65} {\hspace{0.1cm} GT Shadows}}
        \adjincludegraphics[width=0.95\textwidth,Clip={0\width} {0.5\height} 
        {0\width} {0\height}]{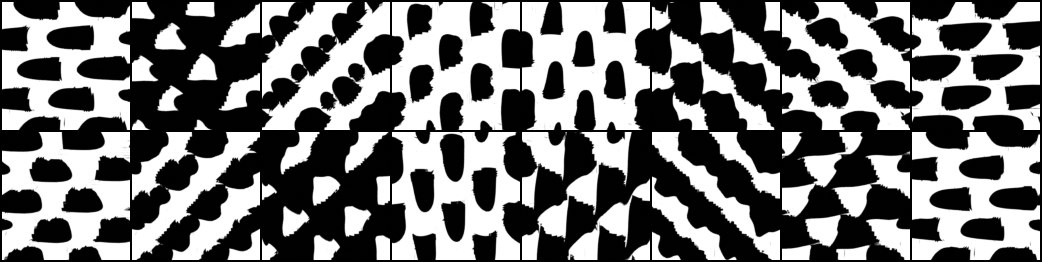}
    \end{tabular}
    
    \begin{tabular}{lc}
        \rotatebox{90}{\scalebox{.65} {\hspace{0.1cm} Est. Shadows}}
        \adjincludegraphics[width=0.95\textwidth,Clip={0\width} {0.5\height} 
        {0\width} {0\height}]{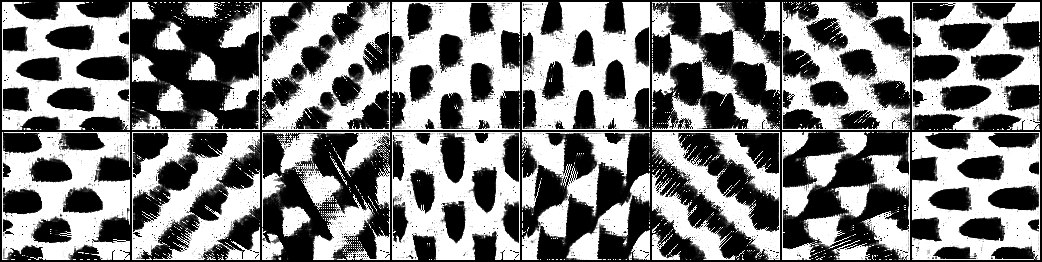}
    \end{tabular}
    
    \begin{tabular}{lc}
        \rotatebox{90}{\scalebox{.65} {\hspace{0.1cm} Shadow Error}}
        \adjincludegraphics[width=0.95\textwidth,Clip={0\width} {0.5\height} 
        {0\width} {0\height}]{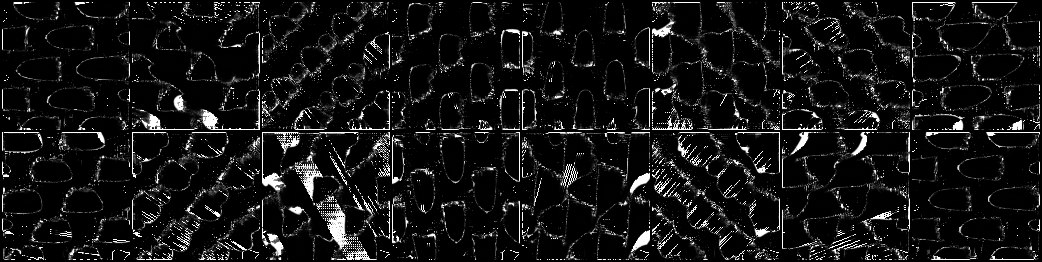}
    \end{tabular}    
    
    \caption{Shadow reconstruction error. The top 3 rows are from the \textit{cactus} object
    and the bottom 3 rows are from the \textit{surface} object. Ground-truth and 
    estimated shadows are shown, along with the L1 error between them. 
    Each column represents a different illumination direction.} 
    \label{fig:shadow_reconstruction}
\end{figure}

We present the shadow reconstruction error on two objects from our rendered dataset.
The reconstruction has two types of errors.
The first is due to the nearest neighbor rounding used for the boundary sampling in the
R2 method, and can be seen clearly in the third row \textit{cactus'} shadow error, in the upper area, and
in the \textit{surface's} second, third and sixth rows.
The second error can mostly be seen in the edges of the objects (e.g., last row in
\Cref{fig:shadow_reconstruction}). The source of the error are edges in the depth map.
Recall we generate a shadow line scan from the light source to each boundary pixel, and estimate the depth
map value for every pixel in the image. 
The error can be minimized by sampling each line in a denser fashion rather than 
sampling a coordinate for every pixel, although it would come at a cost
of computational cost.

\subsection{Results on objects from Dome dataset}
\begin{figure}[t] \centering
    \makebox[0.15\textwidth]{\scriptsize Light 1}
    \makebox[0.15\textwidth]{\scriptsize Light 2}
    \makebox[0.15\textwidth]{\scriptsize Light 3}
    \makebox[0.15\textwidth]{\scriptsize Light 4}
    \makebox[0.15\textwidth]{\scriptsize Light 5}
    \makebox[0.15\textwidth]{\scriptsize Light 6}
    \\
    \begin{tabular}{lc}
        \rotatebox{90}{\scalebox{.7} {\hspace{0.1cm} Input Images}}
        \includegraphics[width=0.95\textwidth]{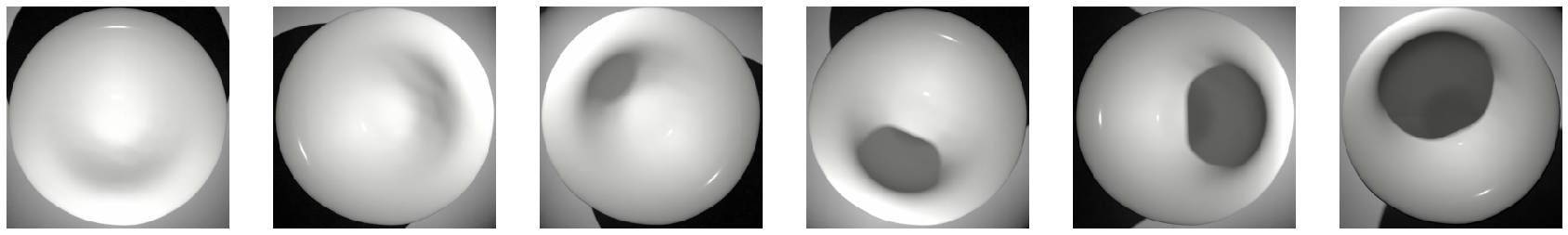}
    \end{tabular}
    
    \begin{tabular}{lc}
        \rotatebox{90}{\scalebox{.7} {\hspace{0.1cm} Est. Shadows}}
        \includegraphics[width=0.95\textwidth]{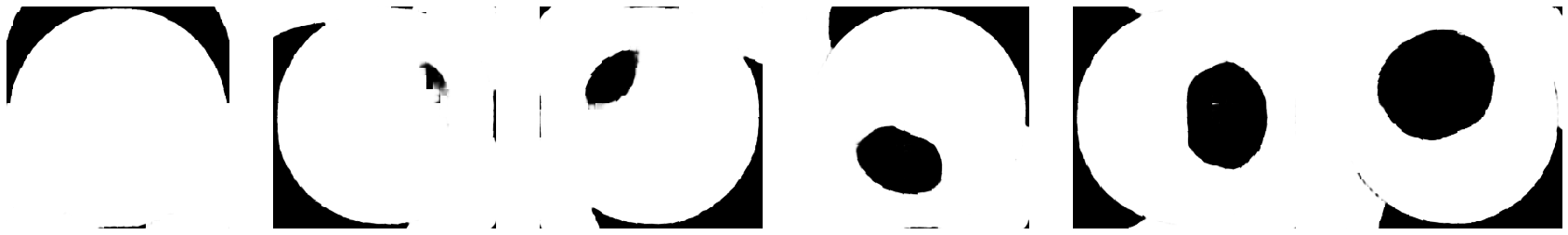}
    \end{tabular}
    
    \begin{tabular}{lc}
        \rotatebox{90}{\scalebox{.7} {\hspace{0.1cm}  Threshold=0.4}}
        \includegraphics[width=0.95\textwidth]{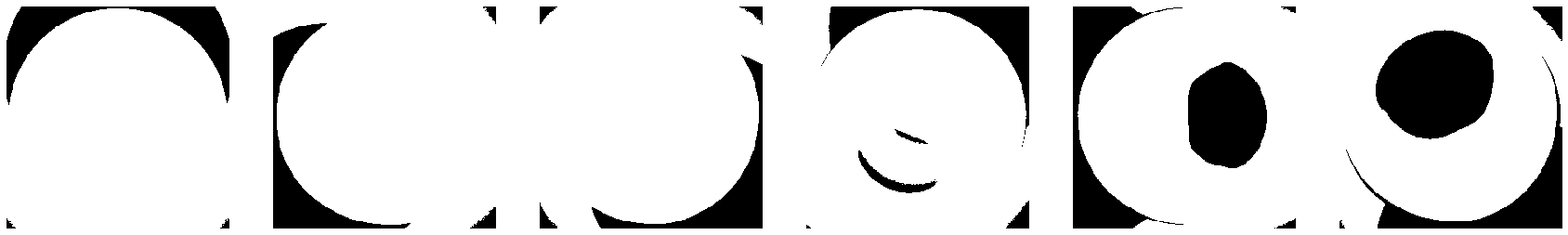}
    \end{tabular}
    
    \begin{tabular}{lc}
        \rotatebox{90}{\scalebox{.7} {\hspace{0.1cm}  Threshold=0.5}}
        \includegraphics[width=0.95\textwidth]{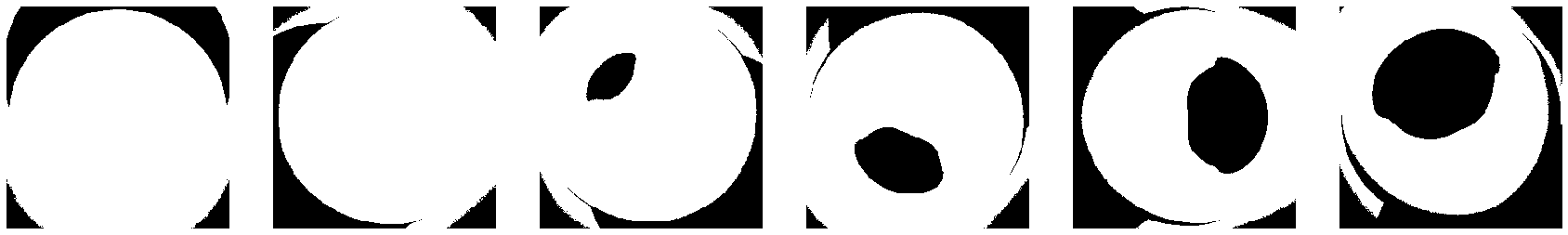}
    \end{tabular}
    \begin{tabular}{lc}
        \rotatebox{90}{\scalebox{.7} {\hspace{0.1cm}  Threshold=0.6}}
        \includegraphics[width=0.95\textwidth]{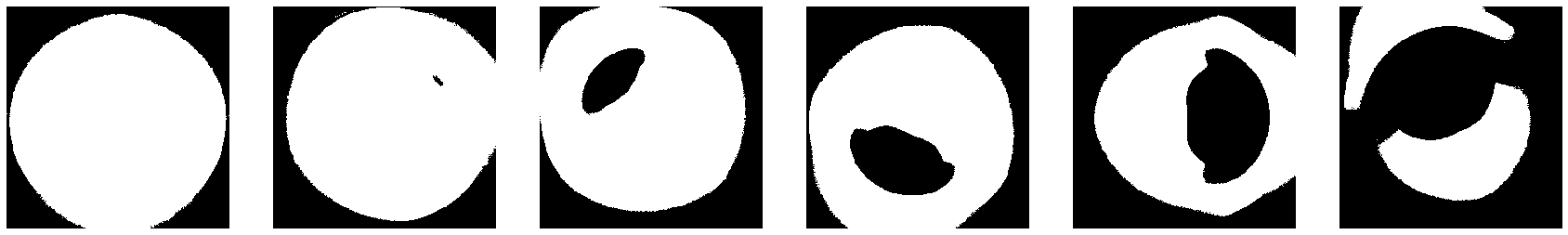}
    \end{tabular}

    \caption{Vase object shadow estimations. 
    The top row contains the input images, the second row is our estimated
    shadow results using the model described in \Cref{sec:shadow-net}. 
    The three bottom rows are a baseline result by taking thresholds of 0.4, 0.5 and
    0.6 over the grayscale levels. Each column is a different light direction. 
    A threshold of 0.4 fails on light directions 2, 3, and 4; a threshold of 0.5 fails on 
    light directions 2 and 6 (external region); a threshold of 0.6 fails on light directions
    4, 5, and 6.} 
    \label{fig:vase_results}
\end{figure}

We present our results on \textit{vase}, \textit{face} and \textit{golf ball} objects from the Dome dataset 
\cite{kaya2021uncalibrated}. The \textit{vase} results can be seen in \Cref{fig:vase_results}.
Previous shape-from-shadow methods \cite{shape_from_shadows,shadow_carving,shadow_graphs} have used a 
threshold on gray-scale images to estimate shadows from images.
As can be seen in \Cref{fig:vase_results}, simple thresholds fail on an object 
such as the \textit{vase}, which has specular highlights.
We show the results for 3 different thresholds by taking values of 0.4, 0.5 and 0.6.
Each threshold fails to produce accurate shadow maps in specific areas.

\begin{figure}[t] \centering
    \makebox[0.2\textwidth]{\scriptsize GT normals~~~~~}
    \makebox[0.2\textwidth]{\scriptsize Estimated normals~~~~}
    \makebox[0.2\textwidth]{\scriptsize Normal error map~~~~}
    \makebox[0.2\textwidth]{\scriptsize Estimated Depth}
    \\
    \begin{tabular}{lc}
        \rotatebox{90}{\scalebox{.7} {\hspace{1.4cm}Vase}}
        \includegraphics[width=0.2\textwidth]{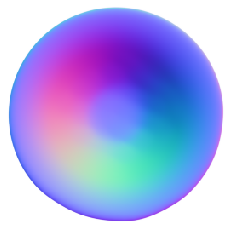}
        \includegraphics[width=0.2\textwidth]{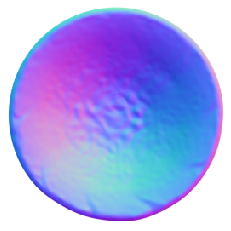}
        \includegraphics[width=0.23\textwidth]{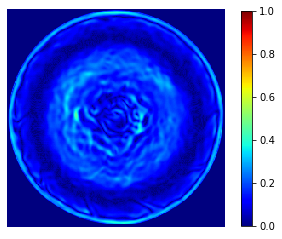}
        \includegraphics[width=0.23\textwidth]{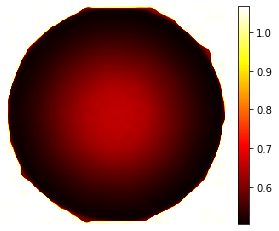}
    \end{tabular}
    
    \begin{tabular}{lc}
        \rotatebox{90}{\scalebox{.7} {\hspace{1.0cm}  Golf ball}}
        \includegraphics[width=0.2\textwidth]{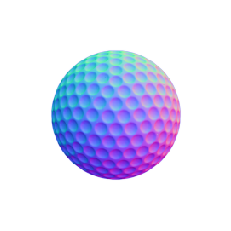}
        \includegraphics[width=0.2\textwidth]{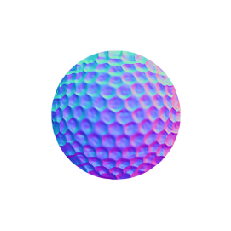}
        \includegraphics[width=0.23\textwidth]{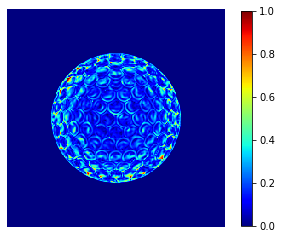}
        \includegraphics[width=0.23\textwidth]{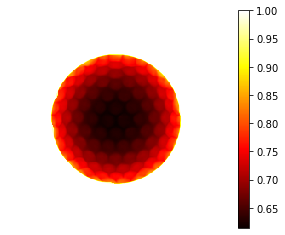}
    \end{tabular}
    
    \begin{tabular}{lc}
        \rotatebox{90}{\scalebox{.7} {\hspace{1.3cm}  Face}}
        \includegraphics[width=0.2\textwidth]{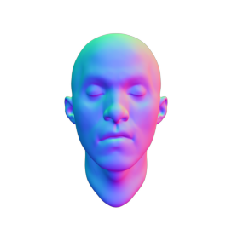}
        \includegraphics[width=0.2\textwidth]{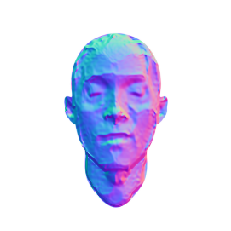}
        \includegraphics[width=0.23\textwidth]{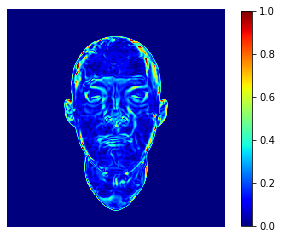}
        \includegraphics[width=0.23\textwidth]{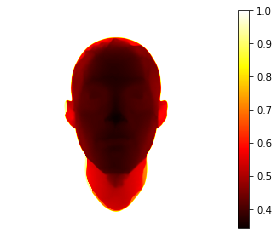}
    \end{tabular}

    \caption{Results on objects from \cite{kaya2021uncalibrated}. 
    Each row (from left to right) consists of ground-truth normals, estimated normals, normal
    error map and estimated depth.
    As described in the paper, the results of DeepShadow on the \textit{vase} object outperform other
    attempted methods. 
    The normal map produced from the \textit{golf ball} has errors around the edges.
    The \textit{face}'s estimated normal map has errors mostly around the forehead area, since that area 
    is smooth and sparse in shadows.}
    \label{fig:dome_data_results}
\end{figure}

We also show our depth and normal estimation results in \Cref{fig:dome_data_results},
which have been discussed in the main paper. 

\subsection{Results on real object}
\begin{figure}[t] \centering
    \begin{tabular}{lc}
    \rotatebox{90}{\scalebox{.7} {\hspace{0.6cm}  Hand Images}}
    \includegraphics[width=0.9\textwidth]{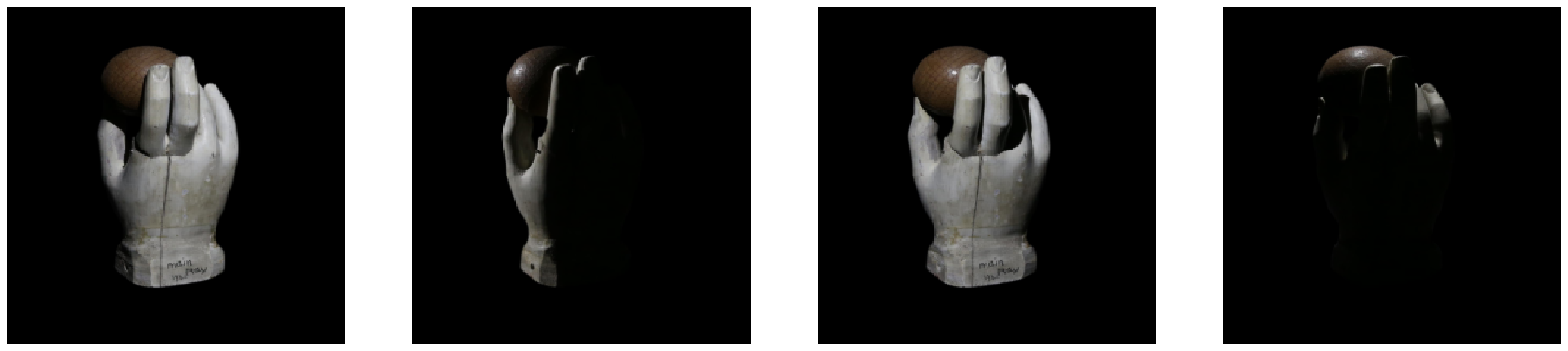}
    \end{tabular}
    \begin{tabular}{lc}
    \rotatebox{90}{\scalebox{.7} {\hspace{0.6cm}  Hand Shadows}}
    \includegraphics[width=0.9\textwidth]{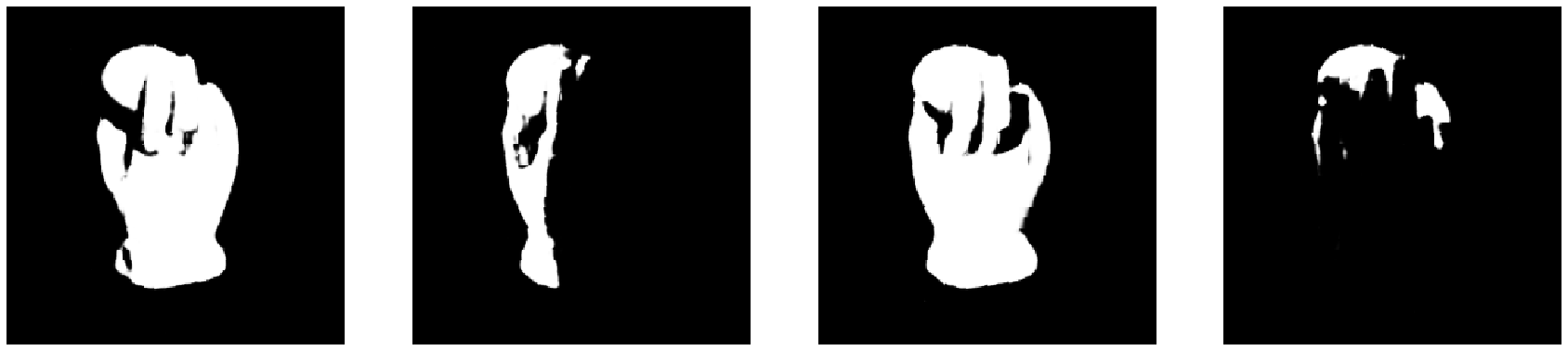}
    \end{tabular}
    \begin{tabular}{lc}
    \rotatebox{90}{\scalebox{.7} {\hspace{0.5cm}  Estimated Results}}
    \includegraphics[width=0.22\textwidth]{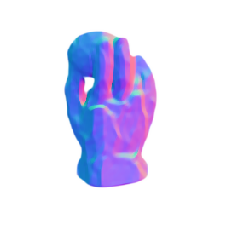}
    \includegraphics[width=0.22\textwidth]{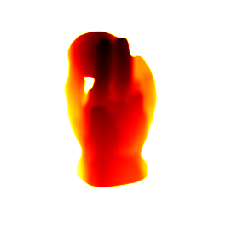}
    \includegraphics[width=0.22\textwidth]{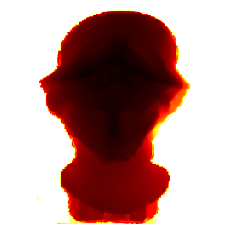}
    \includegraphics[width=0.22\textwidth]{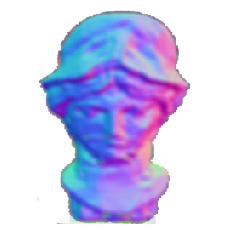}
    \end{tabular}
    \begin{tabular}{lc}
    \rotatebox{90}{\scalebox{.7} {\hspace{0.6cm}  Statue Shadows}}
    \includegraphics[width=0.9\textwidth]{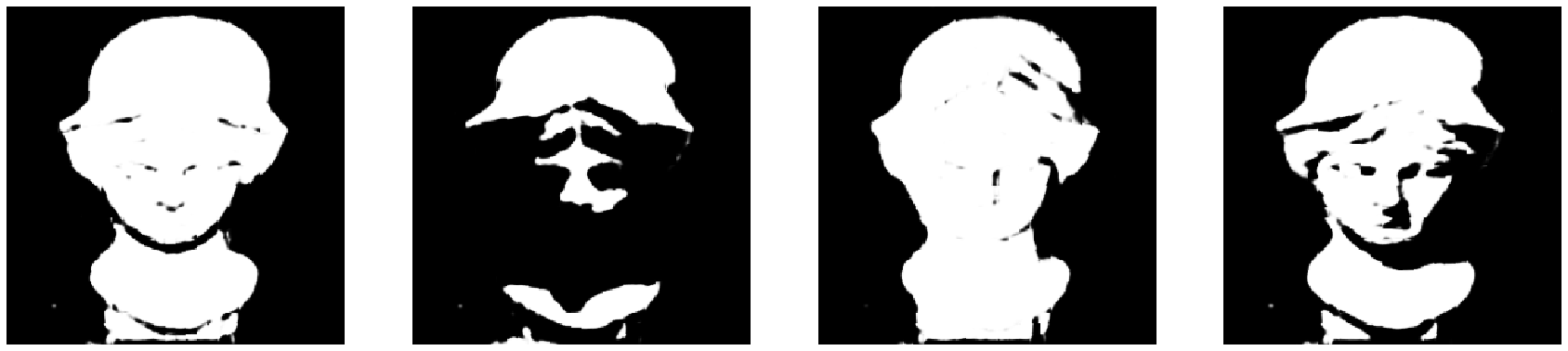}
    \end{tabular}  
    \begin{tabular}{lc}
    \rotatebox{90}{\scalebox{.7} {\hspace{0.6cm}  Statue Images}}
    \includegraphics[width=0.9\textwidth]{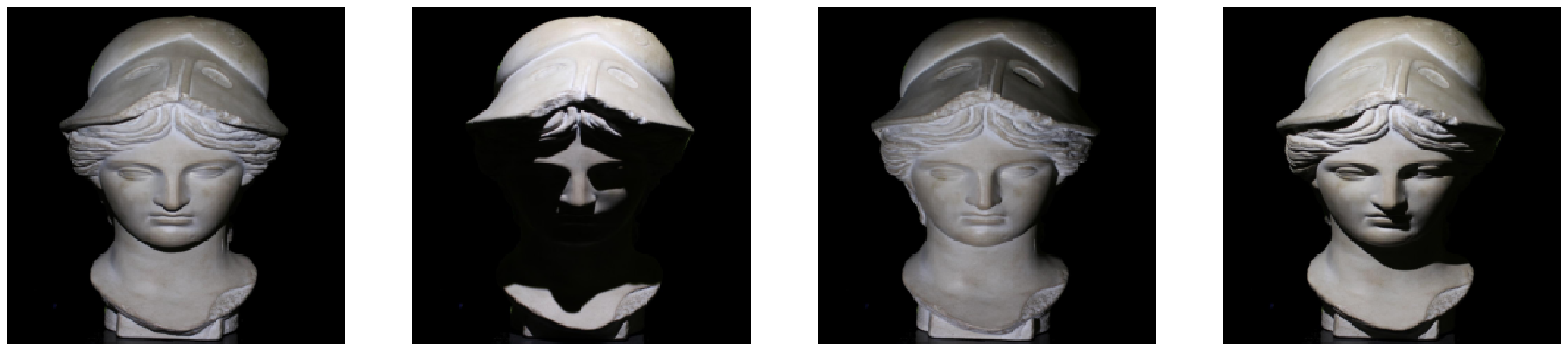}
    \end{tabular}
    
    \caption{Input images, shadows, estimated depth and normals. The upper row includes 
    the \textit{hand} object images, and the second row their estimated shadows.
    The third row contains the estimated depth and surface normals of both
    objects. The fifth row includes the \textit{statue} image inputs and the forth row their estimated
    shadow maps. 
    We can observe DeepShadow is able to extract fine detail in areas such as
    the helmet and hair of the statue, and fails in smooth areas such as its neck.
    }
    \label{fig:real_objects}
\end{figure}

We present qualitative results on the \textit{hand}\footnote{Work of Man Ray, 
sampled at the Museum of Israel.} and \textit{statue}\footnote{Privately sampled.} in
\Cref{fig:real_objects}.
The objects were acquired in a half-dome setting with 34 different illuminations.
Ground truth normals and depth are not available on this dataset, 
as well as light directions, 
which were estimated using the model described in \Cref{sec:shadow-net}.
Our method requires the intrinsic camera parameters which were not known, 
thus had to be roughly estimated by guessing the object's size, 
and assuming a typical 50mm focal length.

\clearpage


\end{document}